\documentclass[conference]{IEEEtran}
\IEEEoverridecommandlockouts

\usepackage{cite}
\usepackage{amsmath,amssymb,amsfonts}
\usepackage{algorithmic}
\usepackage{graphicx}
\usepackage{textcomp}
\usepackage{xcolor}
\usepackage{enumitem}
\usepackage{amsmath, amssymb}
\usepackage{svg}
\usepackage{float}
\usepackage[linesnumbered,ruled,vlined]{algorithm2e}
\SetAlgoSkip{smallskip}
\usepackage{caption}
\usepackage[font=small, skip=2pt]{caption}
\usepackage{subcaption}
\RestyleAlgo{ruled}
\usepackage{multirow}
\usepackage{booktabs}
\usepackage{array}
\SetKw{KwBy}{by}
\def\BibTeX{{\rm B\kern-.05em{\sc i\kern-.025em b}\kern-.08em
    T\kern-.1667em\lower.7ex\hbox{E}\kern-.125emX}}
\begin{document}

\title{CSRAP: Enhanced Canvas Attention Scheduling for Real-Time Mission Critical Perception}

\author{
\IEEEauthorblockN{
Md Iftekharul Islam Sakib\textsuperscript{\dag\S}, 
Yigong Hu{\dag}, Tarek Abdelzaher\textsuperscript{*\dag}
}
\IEEEauthorblockA{
\textsuperscript{\dag}University of Illinois at Urbana-Champaign, 
\textsuperscript{\S}Bangladesh University of Engineering and Technology\\
Email: \{misakib2, yigongh2, zaher\}@illinois.edu
}
}

\maketitle

\begin{abstract}
Real-time perception on edge platforms faces a core challenge: executing high-resolution object detection under stringent latency constraints on limited computing resources. Canvas-based attention scheduling was proposed in earlier work as a mechanism to reduce the resource demands of perception subsystems. It consolidates areas of interest in an input data frame onto a smaller area, called a {\em canvas frame\/}, that can be processed at the requisite frame rate. This paper extends prior canvas-based attention scheduling literature by (i) allowing for {\em variable-size canvas frames\/} and (ii) employing {\em selectable canvas frame rates\/} that may depart from the original data frame rate. We evaluate our solution by running YOLOv11, as the perception module, on an NVIDIA Jetson Orin Nano to inspect video frames from the Waymo Open Dataset. Our results show that the additional degrees of freedom improve the attainable quality/cost trade-offs, thereby allowing for a consistently higher mean average precision (mAP) and recall with respect to the state of the art.

\end{abstract} 


\maketitle
\vspace{-5.5pt}
\section{Introduction}
\label{introduction}
\vspace{-3pt}
Real‑time perception is an increasingly important subsystem of modern mission‑critical platforms, ranging from autonomous cars and aerial reconnaissance vehicles to industrial robotics, all of which require timely processing of high-bandwidth sensor streams under stringent latency and accuracy constraints. Although leading deep‑learning detectors, such as Faster R-CNN~\cite{ren2015towards}, YOLO~\cite{redmon2016you}, and SSD~\cite{liu2016ssd}, exhibit excellent performance at perception tasks, their considerable computational cost remains a significant barrier to deployment in latency-sensitive or resource-constrained scenarios. Consequently, achieving high-quality inference under practical operational constraints necessitates investigating algorithms for judicial allocation of limited computational resources to sensor data inputs across both space (which part of the surveilled environment to consider) and time (how often).

In the context of vision-based applications, real-time perception demands a paradigm shift—from indiscriminate, full-frame inference towards targeted, adaptive scheduling of computational resources that can dynamically prioritize informative regions in a rapidly evolving scene~\cite{liu2020removing}. Particularly challenging are environments characterized by mobile sensing platforms or heterogeneous objects undergoing unpredictable motion, occlusions, or appearance changes. In such settings, the value of assigning perception resources to inspect a given location in the environment inherently depends upon object-level location uncertainty~\cite{liu2022self}, a measure reflecting the ambiguity of an object's state estimation due to elapsed time and dynamic scene conditions.

In this paper, we introduce CSRAP (enhanced Canvas Scheduling for Real-time Attention Prioritization), a novel framework designed to redefine the problem space of real-time adaptive perception scheduling. CSRAP introduces a general, uncertainty-aware canvas-scheduling abstraction that delivers three primary conceptual contributions: (1) an uncertainty-driven optimization objective that dynamically prioritizes objects based on temporal ambiguity and criticality; (2) a flexible canvas-scheduling algorithm that offers new degrees of freedom in canvas-based scheduling, such as a variable canvas frame size and a selectable canvas frame rate; and (3) robust support for mobile or non-stationary camera platforms, enabling real-time adaptation to shifts in viewpoint and environmental complexity.

Previous methods for perception scheduling can be viewed as specialized instances of CSRAP, constrained by narrower assumptions. Specifically, canvas-based attention scheduling methods~\cite{hu2023underprovisioned, hu2024algorithms} optimize computational efficiency by selectively aggregating spatially significant regions into fixed-size canvases for batched processing. While efficient, these approaches neglect temporal uncertainty and rigidly assume a fixed camera viewpoint, limiting their applicability in highly dynamic or mobile scenarios. Additionally, uncertainty-driven methods such as Liu et al.'s self-cueing strategy~\cite{liu2023generalized, liu2022bpb} address temporal ambiguity effectively by dynamically adapting inspection frequency based on predicted object uncertainty growth. While effective in uncertainty mitigation, their reliance on quantized, size-homogeneous batches for GPU inference introduces latency and hampers responsiveness, particularly in scenes with variable object scales or non-uniform motion.

CSRAP subsumes and generalizes these prior approaches, preserving their benefits while eliminating their limitations. An empirical evaluation of realistic video sequences under dynamic sensing conditions demonstrates that CSRAP achieves consistent and substantial gains over state-of-the-art baselines across several metrics. By eliminating unnecessary full-frame inspections and adapting in real-time to object motion and scene complexity, CSRAP significantly reduces both computational overhead and object-level uncertainty. Moreover, its support for multi-resolution canvas scheduling and mobile sensing contexts makes it broadly applicable across diverse edge deployment scenarios, delivering scalable, low-latency perception without sacrificing accuracy.

The rest of the paper is organized as follows. Sections~\ref{sec:system_overview} and~\ref{sec:scheduling-algo} describe canvas scheduling preliminaries and the core innovations underpinning CSRAP, respectively. Section~\ref{sec:evaluation} presents a comprehensive evaluation validating CSRAP's effectiveness across key performance metrics. Section~\ref{sec:related_work} reviews related work. Finally, Section~\ref{sec:conclusion} summarizes the key contributions of this study and future research directions.
\section{Canvas Scheduling Preliminaries: System Overview and Problem Formulation}
\label{sec:system_overview}

\begin{figure}[htb!]
\begin{center}
  \includegraphics[width=8.9cm]{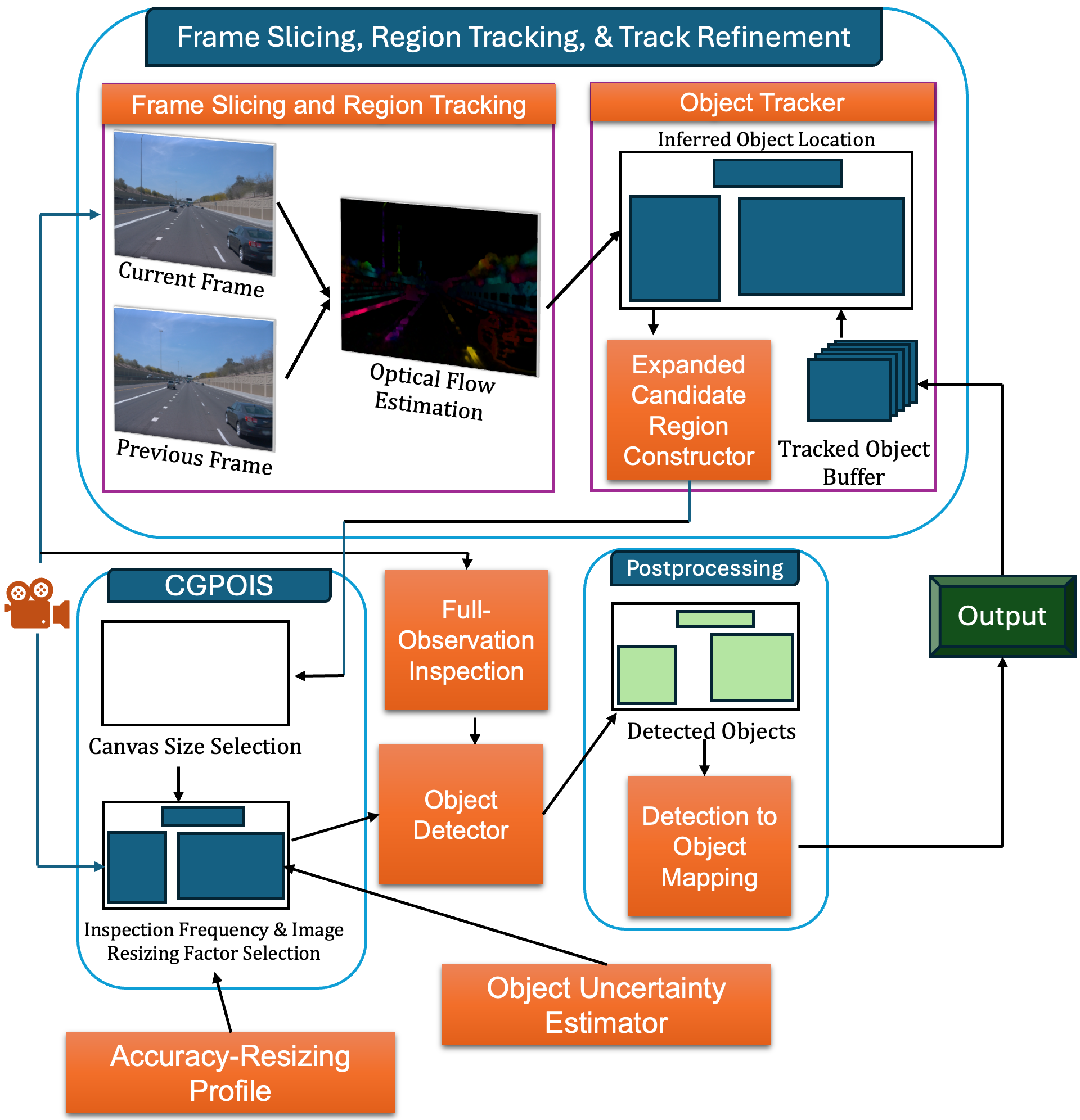}
   \caption{A System Architecture for Canvas-based Attention Scheduling}
   \label{fig:csrpsa}
\end{center}
\end{figure}

This section reviews general canvas scheduling (shown in Figure \ref{fig:csrpsa}) before delving into CSRAP extensions. We assume an IoT device that uses a complex perception sensor, such as a camera or a LiDAR, to continuously observe the surrounding environment. The sensor produces a continuous stream of observations (or frames) $F$ at a constant rate, where each observation has a volume $V$. We use the term ``volume" here for generality; in the case of a camera sensor, it specifically refers to the area of the two-dimensional frames. In this work, we will use the terms ``volume" and ``area" interchangeably, as we are focusing on a camera sensor.

The IoT device uses an onboard perception system to process the observations and make semantically meaningful inferences based on application needs. For example, in a traffic surveillance scenario, a deep learning-based object detector like YOLO may be used to localize and classify objects (e.g., pedestrians, vehicles, bicycles) in camera frames. While the IoT device has an accelerator to process these observations, the accelerator lacks sufficient computational power to process entire frames in real time. The accelerator ( a GPU in our implementation) has a bounded processing capacity, denoted by $V_{GPU}$, such that $V_{GPU} < V$. This general problem setting is consistent with prior work~\cite{hu2023underprovisioned, hu2024algorithms}. 

\setlength{\parskip}{0pt}
To process observations in a timely manner with limited computing resources, our system avoids processing every incoming frame in full.   Instead, similarly to the approach used in self-cueing attention scheduling systems~\cite{liu2023generalized, liu2022self}, the system performs periodic full inspections of observations—termed full-observation inspections—at a defined interval, such as once every ten observations. Between consecutive full-observation inspections, we identify the approximate location of objects of interest (Let's denote this set of objects as $O_k$ at time k) detected in the previous frame by much faster algorithms like optical flow~\cite{KroegerTDG16} and intermittently inspect these regions for exact object localization using adaptive image resizing and canvas-based scheduling to make the most efficient use of limited computational resources. We refer to this integrated approach—combining intermittent partial inspections, canvas scheduling, and image resizing—as canvas-based generalized partial-observation inspection. Since objects are not continuously inspected, their predicted locations may gradually diverge from their actual positions over time. This divergence is referred to as object location uncertainty or object uncertainty in short. We call the interval between two successive full-observation inspections a {\em scheduling horizon\/}. Following the criterion used in prior work on attention scheduling~\cite{liu2023generalized, liu2022self}, our objective for the scheduling algorithm is to minimize the highest weighted uncertainty across all tracked objects, taking into account both the importance of each object and the dynamics of its movement. By means of this intelligent scheduling mechanism, the system maintains high perceptual accuracy while operating within strict computational constraints.

CSRAP, as described above, consists of four key components: (i) Frame Slicing, Region Tracking, and Track Refinement, (ii) Image Resizing and Accuracy Degradation, (iii) Object Uncertainty Estimation, and (iv) Canvas-based Generalized Partial-Observation Inspection. Of these, the last component constitutes our primary contribution. The first three are covered as preliminaries to make the overall description self-contained. Figure \ref{fig:csrpsa} provides an overview of the system architecture, highlighting how each component contributes to the overall scheduling and perception workflow.

\subsection{Frame Slicing, Region Tracking, and Track Refinement}
The design of this module follows prior work on identifying objects in frames in self-cueing attention scheduling systems~\cite{liu2022self,liu2023generalized}. It is responsible for (i) predicting object locations in frames not selected for inspection, and (ii) estimating positional uncertainty for scheduling. We briefly review the core procedures below.

\subsubsection{Frame Slicing and Region Tracking}

Let $O_k = \{o_1, \dots, o_N\}$ denote objects detected at frame $k$ through full-frame inspection, each with bounding box:
\[
B_i^{(k)} = [x_{\min}^{(k)}, y_{\min}^{(k)}, x_{\max}^{(k)}, y_{\max}^{(k)}]
\]

In subsequent frames within the scheduling horizon, object locations are predicted using optical flow based on the most recent observation (partial or full). Let $F_{k \rightarrow k+1}$ denote the optical flow field between frames $k$ and $k+1$. For each object $o_i$, with bounding box $B_i^{(k)}$ in frame $k$, the displacement is estimated by computing the median flow vector\footnote{The median is chosen to reduce the influence of outliers and improve robustness.} over all pixels within $B_i^{(k)}$. The predicted location in frame $k+1$ is then updated as:

\[
\bar{B}_i^{(k+1)} = B_i^{(k)} + \text{median}(F_{k \rightarrow k+1}(B_i^{(k)}))
\]

To account for flow noise and motion uncertainty, we extend this estimate to a conservative \textit{expanded region} based on the extrema of pixel-wise displacements within the previously expanded region $\hat{B}_i^{(k)}$:
\[
\hat{B}_i^{(k+1)} =
\left[
\begin{aligned}
&\hat{x}_{\min}^{(k)} + \min(d_{\hat{x}_i}),\quad \hat{y}_{\min}^{(k)} + \min(d_{\hat{y}_i}), \\
&\hat{x}_{\max}^{(k)} + \max(d_{\hat{x}_i}),\quad \hat{y}_{\max}^{(k)} + \max(d_{\hat{y}_i})
\end{aligned}
\right]
\]
where $d_{\hat{x}_i}$ and $d_{\hat{y}_i}$ are the horizontal and vertical displacement flow components within $\hat{B}_i^{(k)}$.

\subsubsection{Track Refinement via Detection Association}

To maintain accurate object trajectories over time, existing tracks are periodically refined by associating them with new detections. Each active track is represented by its most recent inferred location. The association is performed using the Hungarian algorithm, where the cost between a track and detection is computed via the Intersection-over-Union (IoU) metric, capturing their spatial overlap.

\subsection{Image Resizing and Accuracy Degradation}

To improve computational efficiency, the system uses image resizing as a tunable trade-off. Instead of processing objects at their native resolution, bounding boxes are resized before being packed into the canvas, reducing inference costs at the expense of potential accuracy degradation. For each object $O_i$ with size $S_i$, a resizing ratio $r_j \in (0, 1]$ is applied, where $r_j = 1$ implies no scaling, and smaller values correspond to more aggressive downsampling. The relationship between $r_j$ and detection accuracy is modeled by a pre-profiled degradation function $A_{S_i}(r_j)$, which estimates the expected accuracy at the resized resolution. Empirically, $A_{S_i}(r_j)$ is monotonically increasing and concave, with larger objects exhibiting less sensitivity to downscaling than smaller ones.

\subsection{Estimation of Object Uncertainty}

We adopt the uncertainty formulation in~\cite{liu2023generalized, liu2022self}, modeling positional confidence degradation over time and resolution. The uncertainty $\mathcal{U}_i(t)$ for object $O_i$ at time $t$ is defined as:
\[
\mathcal{U}_i(t) = \frac{w_i (t - t_i)}{A_{S_i}(r_j)}
\]
Where:
\begin{itemize}
    \item $t_i$ is the start time of the most recent inspection on $O_i$;
    \item $A_{S_i}(r_j)$ captures accuracy loss at resized scale;
    \item $w_i = v_i \cdot u_i$ is a weighted product of application-level importance $v_i$ and object-specific uncertainty growth $u_i$. The uncertainty growth rate $u_i$ reflects spatial divergence over the scheduling horizon. We assume $u_i$ is fixed in a scheduling horizon and calculate it as follows:
\[
u_i = \sqrt{\frac{S_i^{\text{ECR}}}{S_i^f}} \cdot \frac{1}{t_f}
\]
Where:
\begin{itemize}
    \item $S_i^{\text{ECR}}$ is the area of the expanded candidate region of $O_i$ from the initial partial observation within the horizon,
    \item $S_i^f$ is the bounding box area of $O_i$ from the last full-frame detection within the scheduling horizon,
    \item $t_f$ is the latency associated with that full inspection.
\end{itemize}
\end{itemize}

Higher values of $\mathcal{U}_i(t)$ indicate lower confidence in the inferred object location, prompting prioritized re-inspection.

\subsection{Canvas-Based Generalized Partial-Observation Inspection}

The Canvas-Based Generalized Partial-Observation Inspection module is the new component of CSRAP's perception scheduling framework and is a departure from prior work on canvas-based scheduling~\cite{hu2023underprovisioned, hu2024algorithms}, where canvas size and period were generally fixed. Instead, the objective of our canvas-based scheduling algorithm includes the selection of canvas size while organizing object inspections through a set of canvases within each scheduling horizon. By carefully selecting which objects to inspect, how to resize them, which canvas size to select based on the current system load, and how to pack the resized objects in the selected-size canvas, the system aims to minimize the highest weighted location uncertainty among all tracked objects.

At the core of this module is the concept of canvas-based scheduling, where multiple objects are spatially arranged and packed into a canvas—a shared image buffer—for processing in one pass. Rather than inspecting the full observation frame or processing each object individually with or without batch, CSRAP resizes selected object regions and places them within a common canvas. The entire canvas is then processed in a single inference pass, enabling the system to amortize computational cost across multiple objects. This strategy significantly improves efficiency and enables frequent inspection of multiple objects without exceeding processing capacity.

The core idea behind this module is that increasing the number of processed canvases within a scheduling horizon allows for more frequent inspections of objects, thus reducing overall uncertainty. However, due to resource constraints, the system must carefully decide not only how much to resize objects but also what canvas size to select for processing. This introduces a nuanced, two-dimensional trade-off:

\begin{itemize} 
    \item \textbf{Object Resizing ($r_j$):}
    Smaller resizing ratios ($r_j$) allow more objects to be packed onto a single canvas, increasing inspection frequency and thereby reducing temporal uncertainty (i.e., lowering $t - t_i$). However, this comes at the cost of reduced detection accuracy (lower $A_{S_i}(r_j)$), increasing perceptual uncertainty. In contrast, larger $r_j$ values preserve accuracy by maintaining resolution but constraining the number of objects per canvas, reducing inspection frequency and increasing temporal uncertainty.

    \item \textbf{Canvas Size:} Larger canvases enable higher-resolution placement (larger $r_j$), improving detection accuracy and lowering perceptual uncertainty. However, due to computational constraints, fewer large canvases can be processed per scheduling horizon, decreasing inspection frequency and increasing temporal uncertainty. On the other hand, the smaller canvases can be processed more frequently within the same compute budget, enhancing inspection frequency but requiring more aggressive resizing, which amplifies perceptual uncertainty.
\end{itemize}

This interplay between canvas size selection, object resizing, and canvas allocation requires careful scheduling to ensure an optimal balance between inspection frequency (temporal uncertainty) and image accuracy (perceptual uncertainty) while staying within computational constraints. This module aims to optimally manage this trade-off to minimize the maximum weighted uncertainty across all tracked objects within resource constraints.

\subsubsection{Problem Formulation}

Let us consider a system operating under a constrained computational resource budget denoted by $R$. The system supports a finite set of candidate canvas sizes, represented as $C_{g} = \{C_1,\dots, C_n\}$, where each $C_i$ corresponds to a canvas with spatial capacity $S_{C_i}$ (e.g., $512 \times 512$ pixels). With the limited resource $R$, the number of size-$C_i$ canvases processable within a horizon is:
$$
C_{f} = \left\{\left\lfloor \frac{R}{C_1} \right\rfloor, \left\lfloor \frac{R}{C_2} \right\rfloor, \dots, \left\lfloor \frac{R}{C_n} \right\rfloor \right\}
$$

For each selected canvas size $C_s \in C_g$, the scheduler determines a subset of objects $O_s \subseteq O_k$—where $O_k$ denotes all currently tracked objects—to be placed within the canvas, along with resizing ratios $r_j \in (0,1]$ for each $o_i \in O_s$. The primary goal of this scheduling module is to jointly select $C_s$, $O_s$, and $\{r_j\}$ to minimize the maximum uncertainty across all tracked objects, i.e., $\min_{\{C_s, O_s, r_s\}} \max_{o_i \in O_k} \mathcal{U}_i(t)$, subject to the following constraints:

\begin{itemize}
    \item \textbf{Canvas packing constraint:} The total area occupied by all resized objects within a single canvas must not exceed its spatial capacity $S_{C_{i}}$:
    $$
    \sum_{o_i \in O_s} \text{Area}(o_i, r_j) \leq S_{C_{i}}
    $$

    Note that $S_{C_i}$ may be less than the full canvas area, depending on the packing heuristic employed. For instance, if the heuristic guarantees successful packing only when the total object area is less than half the canvas area, then $S_{C_i}$ is effectively bounded by $0.5 \times \text{Area}(C_i)$.

    \item \textbf{Canvas Budget Constraint:} For any canvas size $C_i$, the number of canvases processed within the scheduling horizon must satisfy:
    $$
    N_{C_i} \leq \left\lfloor \frac{R}{C_i} \right\rfloor
    $$
   
    Where $N_{C_i}$ denotes the number of canvases of size $S_{C_i}$ allocated within the current scheduling horizon, and $\left\lfloor \frac{R}{C_i} \right\rfloor$ is the maximum number of such canvases that can be processed under the resource budget $R$.

    \item \textbf{Canvas Selection Constraint:} Each scheduling horizon consists of $H_l$ frames, where the first frame is inspected in full resolution, and the remaining $H_l - 1$ frames rely on canvas-based inspection. Hence, only canvas sizes $C_i$ satisfying the following viability condition can be chosen:

    $$
    1\leq \left\lfloor \frac{R}{C_{\text{i}}} \right\rfloor
    $$
\end{itemize}

This formulation clearly defines the trade-offs and constraints involved, enabling the scheduling module to balance object inspection frequency, perceptual accuracy, and computational constraints effectively.

To better illustrate the problem, let's consider a traffic surveillance system tasked with tracking objects. At time $t$, the system observes a pedestrian ($o_1$) positioned close to the camera, appearing relatively large in the frame; two vehicles ($o_2$, $o_3$); two bicycles ($o_4$, $o_5$); and two distant traffic signs ($o_6$, $o_7$) occupying smaller pixel areas due to their distance from the camera. At full resolution ($r_j = 1$), the objects occupy the following approximate areas: $o_1$ (pedestrian): $128 \times 128$ pixels; $o_2$, $o_3$ (vehicles): $96 \times 96$ pixels each; $o_4$, $o_5$ (bicycles): $64 \times 64$ pixels each; and $o_6$, $o_7$ (traffic signs): $48 \times 48$ pixels each. Based on application-specific priorities, the pedestrian ($o_1$) is most important. The object importance hierarchy is as follows:
$$
w_{o_1} > \{w_{o_2}, w_{o_3}\} > \{w_{o_4}, w_{o_5}\} > \{w_{o_6}, w_{o_7}\}
$$

Let's assume a scheduling horizon $H_l = 4$, where the first frame is inspected at full resolution, and the remaining $H_l - 1 = 3$ frames are processed using partial-frame inspections. The system considers a candidate set of canvas sizes $C_g = \{C_1, C_2, C_3\}$, with spatial capacities: $S_{C_1} = 128 \times 128$, $S_{C_2} = 256 \times 256$, and $S_{C_3} = 512 \times 512$. Given the computational budget $R$, the system can process up to three $C_1$ canvases, two $C_2$ canvases, or one $C_3$ canvas during the partial-frame inspection slots. All candidate sizes satisfy the Canvas Selection constraint and are thus feasible for use.

The decision of which canvas size to select involves balancing temporal and perceptual uncertainty. For example, selecting $C_1$ (smallest canvas) allows three inspections per horizon, improving temporal coverage but requiring aggressive object resizing. The pedestrian $o_1$ must be downscaled to at least $r_{o_1} = 0.3$ to fit within a $128 \times 128$ canvas, thereby increasing perceptual uncertainty due to reduced resolution.

Conversely, selecting the largest canvas $C_3 = 512 \times 512$ enables full-resolution processing for all objects, eliminating perceptual uncertainty from resizing. However, only one canvas can be processed per horizon, increasing temporal uncertainty due to reduced inspection frequency.

Ultimately, the system must intelligently balance these competing sources of uncertainty. The optimal configuration selects a canvas size and associated resizing strategy that minimizes the maximum object uncertainty $\mathcal{U}_i(t)$ across all tracked objects, subject to the resource constraint $R$.
\section{Canvas-based Generalized Partial-Observation Inspection Scheduling Algorithm}
\label{sec:scheduling-algo}

The Canvas-Based Generalized Partial-Observation Inspection Scheduling problem aims to determine the optimal canvas size, select a set of objects along with their inspection frequencies, and choose appropriate resizing decisions to efficiently pack these objects into a limited number of optimal size canvases. The main objective is to minimize the maximum uncertainty across all tracked objects.

To understand the complexity of this problem, let us consider a simplified version of this problem where the optimal canvas size is already determined, and all tracked objects have the same level of uncertainty (thereby, they require the same inspection frequency). Even in this simplified scenario, the problem turns into a variant of the multi-choice multi-bin packing problem, which is known to be NP-hard. Even if we simplify it further to the basic task of packing a fixed number of objects of different sizes into a fixed-size canvas, we encounter the classic 2D bin packing problem, which is also NP-hard. Since both of these simplified variants are computationally NP-hard, it logically follows that the original canvas-based Generalized Partial-Observation Inspection Scheduling problem is NP-hard as well. Given this inherent complexity, seeking exact solutions in real-time settings with limited computational resources is not feasible. Hence, we need to develop an efficient and near-optimal greedy approximation algorithm that is feasible for resource-constraint real-time scheduling.

\begin{algorithm}[ht]
\caption{The CGPOIS Policy}
\label{alg:CGPOIS} 
\KwIn{Object set $O = \{O_1, \dots, O_n\}$, Weights $W = \{w_1, \dots, w_n\}$, Set of allowed canvas sizes $\{C_1, \dots, C_m\}$, Accuracy matrix $A = [A_{sj}]$, where $A_{sj}$ is the expected accuracy for size $s$ at resizing ratio $r_j$, Scheduling horizon length $H_l$, Available computational resource $R$ for processing $H_l-1$ frames, Allowable resizing factor set $\mathcal{R}_f$, Minimum accuracy threshold $Ac_{min}$, Maximum accuracy threshold $Ac_{max}$, minimum inspection frequency $I_{min}$}
\KwOut{An optimal feasible schedule with minimized uncertainty}

Sort and re-index $W$ such that $w_1 \le \cdots \le w_n$\;
$I \gets \emptyset$\;
\For{$i \gets 1$ \KwTo $n$}{
    $I \gets I \cup \lfloor I_{min} + \left( \frac{w_i - w_1}{w_n - w_1} \right) \cdot (H_l -I_{min}-1) \rfloor$

}

$C_s \gets \emptyset$\;
\For{$i \gets 1$ \KwTo $m$}{
    \If{$1 \leq \left\lfloor \dfrac{R}{C_i} \right\rfloor \leq H_l - 1$}{
        $C_s \gets C_s \cup \{C_i\}$\;
    }
}
$\mathcal{U}_{\text{min}} \gets \infty$\;
$R_s \gets \emptyset$\;
\ForEach{$C_i \in C_s$}{
    Call Algorithm~\ref{alg:casr} with $O$, $I$, $S_{C_i}$, $A$, $R$, $\mathcal{R}_f$, $Ac_{min}$, $Ac_{max}$\;
    Let $\mathcal{U}_i$ be the uncertainty returned by Algorithm~\ref{alg:casr}\;
    \If{$\mathcal{U}_i < \mathcal{U}_{\text{min}}$}{
        $\mathcal{U}_{\text{min}} \gets \mathcal{U}_i$\;
        $\mathcal{S}_{opt} \gets$ Schedule returned by Algorithm~\ref{alg:casr}\;
    }
}
\Return{$\mathcal{S}_{opt}$}\;
\end{algorithm}

\begin{algorithm}[ht]
\caption{Canvas-based Partial-observations Inspection Scheduling (CPOIS)}
\label{alg:casr}
\KwIn{Object set $O = \{O_1, \dots, O_n\}$, Inspection frequency set $I = \{I_1, \dots, I_n\}$, Canvas Size $S_{c}$, Accuracy matrix $A$, Available computational resource $R$, Allowable resizing factor set $\mathcal{R}_f$, Minimum Accuracy Threshold $Ac_{min}$, Maximum Accuracy Threshold $Ac_{max}$}
\KwOut{An optimal schedule for the inspection frequency set, accompanied by the corresponding maximum object-level uncertainty}

$C_{count} \gets \left\lfloor \dfrac{R}{S_C} \right\rfloor$\;

Assume a set of $C_{\text{count}}$ canvases $\{1,\ldots,C_{count}\}$, each with size $S_C$ and an initial load of 0, i.e., $\text{load}(C_j) = 0$ for all $j \in \{1, 2, \dots, C_{\text{count}}\}$\;

Sort and re-index $I$ such that $I_1 \ge \cdots \ge I_n$\;

Find the nonzero 
$ \{I_1, I_2, \dots, I_h \mid I_i \geq 1 \}$

\For{$i \gets I_1, I_2, \dots, I_h$}{
    Let $C_c$ be the first $\lfloor \frac{C_{count}}{i} \rfloor$ canvases, i.e., $\{C_1, \dots, C_{\lfloor \frac{C_{count}}{i} \rfloor}\}$\;
    
    Place the first inspection task of object $O_i$ onto a canvas in $C_c$ with the lowest current load, then increase that canvas's load by $s_i$\;

    Replicate this mapping for the remaining $i-1$ inspection tasks of the same object $O_i$ to the remaining subset of canvases\;
}


Call Algorithm~\ref{alg:cascmap} with $O$, $S_{C_i}$, $A$, $\mathcal{R}_f$, $Ac_{min}$, $Ac_{max}$, $C_{count}$\;
$\mathcal{S} \gets$ Schedule returned by Algorithm~\ref{alg:cascmap}\;

Compute uncertainty set $\mathcal{U} = \{\mathcal{U}_i \mid o_i \in {O}$, given schedule $\mathcal{S}$\}\;

Compute $\mathcal{U}_{\max} = \max_{i} \mathcal{U}_i$\;

\Return schedule $\mathcal{S}$, maximum object uncertainty $\mathcal{U}_{\max}$\;
\end{algorithm}

\begin{algorithm}[ht]
\caption{Feasible Schedule for an Object-Canvas Mapping (FSOCM)}
\label{alg:cascmap}
\KwIn{Object set $O = \{O_1, \dots, O_n\}$, Canvas Size $S_{c}$, Accuracy matrix $A$, Allowable resizing factor set $\mathcal{R}_f$, Minimum Accuracy Threshold $Ac_{min}$, Maximum Accuracy Threshold $Ac_{max}$, Canvas Count $C_{count}$}
\KwOut{An optimal schedule for the object-canvas mapping}

\For{$C \in \{1, \dots, C_{\text{count}}\}$}{
    $t_c \gets$ start of valid period of canvas $C$\;
    
    $F \gets$ most recent camera frame at $t_c$\;
    
    $\mathcal{O}_c \gets$ Objects to be processed in canvas $C$\;
    
    Initialize $\mathcal{A}_c : \{a'_i = 0, \forall o_i \in \mathcal{O}_c\}$, ${U_A} = 0$\;
    
    Compute $G = A^*/E^*$ for all $(O_i, r_j)$ such that $(O_i, r_j) \in \mathcal{O}_C \times \mathcal{R}_f$ and $\text{Ac}_{\text{min}} \le A[S_i, r_j] \le \text{Ac}_{\text{max}}$\;

    Sort and index $G$ such that $G_1 \ge \cdots \ge G_L$\;

    \For{$k = 1,2,\dots,L$}{
        Let $G_k$ belong to object $O_i$ and current selected size $a'_{i_c}$ for the object\;
        
        \If{$U_A + E_{a'_{i_c}}^* \le \frac{S_C}{2}$}{
            $U_A \gets U_A + E_{a'_{i_c}}^*$\;
            $a'_i \gets a'_{i_c}$\;
        }
        \Else{
            $srindex \gets k$\; 
            \textbf{break}\;
        }
    }

    \For{$k' = srindex\dots,L$}{
        Let $G_{k'}$ belong to object $O_i$ and current selected size $a'_{i_c}$ for the object\;
        
        \If{$U_A + E^*_{a'_{i_c}} \le S_C$ \textbf{and} packing valid with $a'_{i_c}$}{
            $U_A \gets U_A + E_{a'_{i_c}}^*$\;
            $a'_i \gets a'_{i_c}$\;
        }
        \Else{
            \textbf{break}\;
        }
    }

    
    Generate packing $\mathcal{P}_c$ from $\mathcal{O}_c$ and $\mathcal{A}_c$\;
    Add $(\mathcal{P}_c, S_c, t_c, F)$ to $\mathcal{S}$\;
}

\Return schedule $\mathcal{S}$\;
\end{algorithm}

To ensure a real-time solution to the considered NP-hard problem, CGPOIS determines an optimal canvas size from a finite set of candidates and allocates objects across multiple canvases of that size. The allocation should be optimized to ensure efficient organization while taking into account the weight of each object and allocating inspection frequency based on the weight. For example, if an object $O_i$ has an inspection frequency of $I_i$, it will be assigned to $I_i$ canvases accordingly. We propose a coordinated pipeline of three algorithms, to achieve the objective of CGPOIS.

The CGPOIS policy, implemented as ALGORITHM~\ref{alg:CGPOIS}, begins by computing inspection frequencies $I = \{ I_1, I_2, \ldots, I_n \}$ for objects $O = \{ O_1, O_2, \ldots, O_n \}$ based on their weight factors $W = \{ w_1, w_2, \ldots, w_n \}$. Initially, the weights are first sorted in ascending order and normalized by the smallest weight. Subsequently, the inspection frequencies are determined by multiplying \( H_l - 1 \) to these normalized weights, ensuring that $I_i \leq H_l - 1$. Next, a subset of permissible canvas sizes $C_1, C_2, \dots, C_s$ is selected from the set of available canvases $C_1, C_2, \dots, C_m$, subject to the canvas selection constraint that limits the number of canvases per horizon to at most $H_l - 1$. For each valid canvas size, an optimal schedule $S_{\text{opt}}$ is generated using the Canvas-based Partial-Observation Inspection Scheduling (CPOIS) algorithm, as presented in ALGORITHM~\ref{alg:casr}. Among all candidates, the schedule with the lowest maximum uncertainty is selected for execution. The final schedule $S$ is represented as an ordered sequence of tuples $(\mathcal{P}^1, s^1, t^1, k^1), (\mathcal{P}^2, s^2, t^2, k^2), \dots, (\mathcal{P}^I, s^I, t^I, k^I)$, where each tuple corresponds to a partial-frame inspection using a canvas. Specifically:

\begin{itemize}[leftmargin=2em]
    \item $\mathcal{P}^i$: a set of resized objects packed into the $i$-th canvas; each object appears at most once within a single canvas.
    \item $s^i$: canvas size used for the $i$-th inspection.
    \item $t^i$: start time of canvas execution.
    \item $k^i \in \{2, \dots, H_l\}$: frame index of the partial-frame inspection.
\end{itemize}

\begin{figure}[t] 
    \centering
    \includegraphics[width=0.5\textwidth]{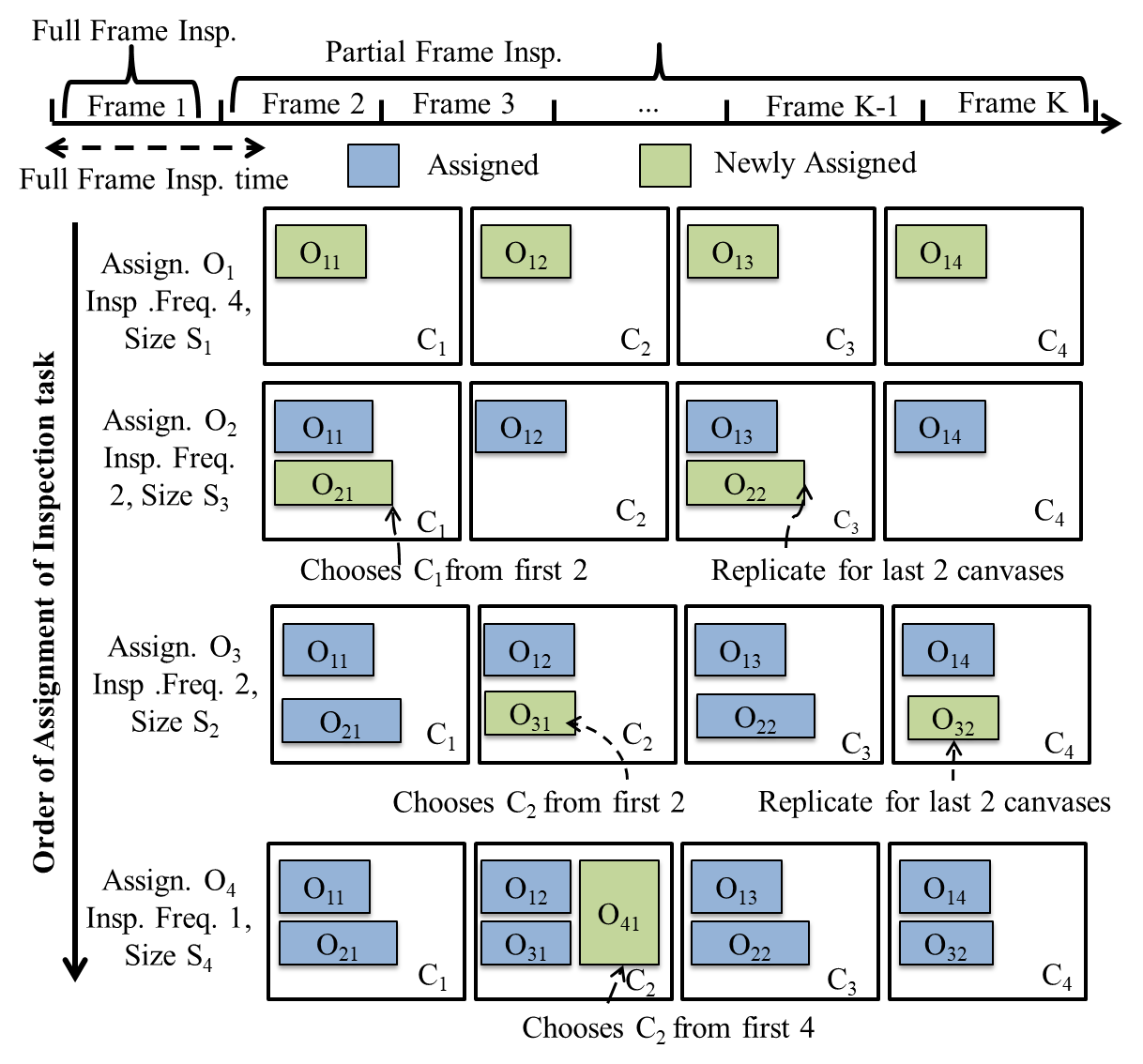}
    \caption{A graphical representation delineates the methodology employed by CPOIS in the generation of the mapping of inspection tasks to canvases.}
    \label{fig:insp_assign} 
\end{figure}

\textbf{ALGORITHM~\ref{alg:casr}} begins by determining $C_{\text{count}}$, the number of canvases of size $S_C$ that can be supported under resource budget $R$. Each canvas $\mathcal{C}_i$ is initialized with zero load: $\text{load}(\mathcal{C}_i) = 0$. Object inspection frequencies $I = \{ I_1, I_2, \ldots, I_n \}$ are sorted in descending order, and only objects with $I_i > 0$ are considered. Each object $O_i$ is then assigned to $I_i$ distinct canvases. The first assignment is made to one of the first $C_{\text{count}}/I_i$ canvases with the minimum load. The remaining $I_i - 1$ replicas are distributed across other canvases to ensure balanced allocation. The assignment process is illustrated in Fig.~\ref{fig:insp_assign}, which shows four objects—represented as tuples $(O_1, 4, S_1)$, $(O_2, 2, S_2)$, $(O_3, 2, S_3)$, and $(O_4, 1, S_4)$—being allocated across four identical canvases $C_1$ through $C_4$. Object $O_1$, requiring four inspections, is mapped to all canvases. $O_2$, $O_3$, and $O_4$ are assigned to canvases with minimum load, resulting in placements on $\{C_1, C_3\}$, $\{C_2, C_4\}$, and $C_2$, respectively.

After completing the object-to-canvas mapping (lines 5–8), \textbf{ALGORITHM~\ref{alg:casr}} invokes the subroutine entitled Feasible Schedule for Object-Canvas Mapping (FSOCM), as detailed in ALGORITHM \ref{alg:cascmap}, to identify the optimal feasible schedule utilizing the best resizing options. Upon obtaining the optimal schedule, the uncertainty of each object is computed. The maximum uncertainty is computed from the set of calculated uncertainties. Ultimately, the final schedule $S$ along with the maximum uncertainty $u_{\text{max}}$ is returned.

The main task of \textbf{ALGORITHM~\ref{alg:cascmap}} is to construct an optimal inspection schedule by selecting appropriate resizing factors that allow efficient canvas utilization without compromising detection accuracy. For each canvas $C$, all assigned objects $\mathcal{O}_C$ are initially allocated zero area (i.e., $a_i' = 0$ for each $O_i \in \mathcal{O}_C$). The algorithm systematically evaluates all combinations $(O_i, r_j) \in \mathcal{O}_C \times \mathcal{R}_f$, where $\mathcal{R}_f$ denotes the set of allowable resizing factors. For each pair, the resized size $S_i$ and corresponding detection accuracy $A[S_i, r_j]$ are computed. Only those configurations satisfying the predefined accuracy bounds $[A_{\min}, A_{\max}]$ are retained. From these candidate configurations, we define a utility score $G_l$ for each pair $(O_i, r_j)$, capturing the efficiency of allocating additional area in exchange for improved accuracy. Each $G_l$ is determined using the following computation:

Let $a_i^{-}$ represent the object size level immediately smaller than the currently selected size $a_i'$, and let $a_i^{+}$ represent the size level immediately larger than $a_i'$. We define the \textit{incremental accuracy gain} as:
\[
A^*_{i, a_i'} = 
\begin{cases}
A(o_i, a_i') - A(o_i, a_i^{-}), & \text{if } a_i' > a_{\min} \\
A(o_i, a_{\min}), & \text{if } a_i' = a_{\min}
\end{cases}
\]

Similarly, the \textit{incremental weighted area} is defined as:
\[
E^*_{a_i'} = 
\begin{cases}
a_i' - a_i^{-}, & \text{if } a_i' > a_{\min} \\
a_i', & \text{if } a_i' = a_{\min}
\end{cases}
\]

The \textit{incremental efficiency}, which measures accuracy gain per unit increase in area, is given by:
\[
G_{i, a_i'} = w_i \cdot \left( \frac{A^*_{i, a_i'}}{E^*_{a_i'}} \right)
\]

After computing all $G_l = G_{i, a_i'}$ for valid object-size pairs, we sort them in descending order of efficiency.
Empirical profiling confirms that $G_l$ typically decreases with increasing object size, reflecting that smaller objects are more sensitive to resolution loss. This monotonic behavior satisfies the Karush-Kuhn-Tucker (KKT) conditions~\cite{kt1}, which lets us apply a greedy selection strategy that is not only efficient but also optimal for our formulation. The resizing process works by repeatedly choosing the highest-ranked $G_l$ values and gradually adjusting each object's assigned size $a_i'$ until the total allocated area comes close to the canvas's capacity limit.

When object and canvas sizes are quantized—as assumed in this work—full canvas utilization is guaranteed due to the discretized geometry, and every selected $G_l$ can be accommodated without overflow~\cite{CoffmanCGJSWY00}. Consequently, no secondary allocation pass is necessary. In contrast, under general (non-quantized) conditions, the algorithm conservatively limits the total allocated area to half the canvas capacity, adhering to established lower bounds for rectangle packing with $90^\circ$ rotation~\cite{pr1}. In such cases, any $(O_i, a_i')$ pair whose corresponding $G_l$ cannot be placed is deferred to a buffer. After the main greedy loop completes, a secondary pass may be performed to attempt packing these buffered items into the remaining space. This secondary step is omitted in our algorithmic implementation, as it is rendered unnecessary under the quantized-size assumption.

After finalizing the object sizes in $A_C$, all objects $\mathcal{O}_C$ are packed into canvas $C$ using their selected resolutions, and this local packing result is integrated into the global schedule $S$. The completed schedule $S$ is then returned. These resizing decisions ensure that partial-frame observations are carried out with maximum perceptual fidelity while strictly adhering to the system's resource constraints.

\section{Evaluation}
\label{sec:evaluation}

We perform a comprehensive set of experiments to evaluate the performance of the proposed Canvas-based Attention Scheduling for Real-Time Mission Critical Perception (CSRAP). This section discusses all the details of the experimental environment, including the dataset and hardware platforms, the baseline schedulers against which CSRAP is compared, and the key performance metrics used for evaluation.

\vspace{-6pt}
\subsection{Experimental Setup}

\subsubsection{Dataset Specification} Waymo Open Dataset~\cite{waymoR} has been employed as the primary benchmark for our evaluation. It provides high-fidelity, multi-sensor recordings from production-grade autonomous vehicles operating in varied urban and suburban contexts. Each sequence spans 20 seconds, captured at 10 Hz with $1920 \times 1280$ resolution. To emulate a forward-facing perception stack common in modern ADAS pipelines, we restrict our analysis to the front-view camera stream.

\vspace{1ex}
\subsubsection{Embedded Hardware Platform} All empirical evaluations are conducted on the NVIDIA Jetson Orin Nano, a low-power embedded platform purpose-built for real-time robotics and autonomous perception. The system features a 6-core Arm Cortex-A78AE CPU, an Ampere-based GPU with 1024 CUDA cores, and 8 GB of unified memory. We configure the device in \texttt{MAXN} mode to enable peak performance across both CPU and GPU subsystems.

\vspace{1ex}
\subsubsection{Detection Backbone}

Our object detection pipeline is built on YOLO11L\cite{yoloR1,yoloR2}, a high-performance single-stage detector implemented in PyTorch. The model is initialized with COCO-pretrained weights\cite{cocoR} and executed in \texttt{FP16} precision to utilize GPU acceleration. Both depth and width multipliers are set to 1.0. Prior to deployment, we profile inference latency across multiple input resolutions to enable resolution-aware scheduling at runtime.

\vspace{1ex}
\subsubsection{Variable Load Emulation}

To emulate variable system load, we introduce controlled variation in frame arrival intervals. A fixed scheduling window of 10 frames is maintained, and the inter-frame delay $P$ is adjusted to one of three values—150 ms, 100 ms, and 70 ms—corresponding respectively to low, medium, and high system pressure. These rates reflect practical operating frequencies of 6.67 Hz, 10 Hz, and 14.3 Hz.

\vspace{1ex}
\subsubsection{Object Criticality}

To model prioritization of perceptual targets, we define object criticality as the product of two interpretable components: (i) a manually assigned \emph{semantic importance score} for each object class, and (ii) an \emph{estimated proximity factor}, approximated using bounding box width under the assumption of class-wise size consistency. Objects exceeding a class-specific bounding box threshold are designated as critical, and we report metrics both for the full object set and this high-importance subset.

\vspace{1ex}
\subsubsection{Evaluation Methodology}

Detection quality is assessed along multiple axes—existence detection, class recognition, and spatial localization. Detections are matched with ground truth using an Intersection-over-Union (IoU) threshold of 0.5. Three key metrics are utilized to help understand precision and recall:

\begin{itemize}
    \item \textbf{Detection Recall (DR):} Fraction of ground-truth objects successfully detected.
    \item \textbf{Detection Precision (DP):} Fraction of predicted objects that are valid matches.
    \item \textbf{Mean Average Precision (mAP):} Aggregated detection+classification metric, computed using a public evaluation tool~\cite{mAPR}.
\end{itemize}

As Liu et al.~\cite{liu2022self} observed negligible variation in localization and classification performance across methods—and our findings concur—we omit these metrics from our reported results.

\subsection{Scheduling algorithm comparison}
\subsubsection{Baseline Scheduling Policies}
We benchmark our proposed scheduler against five representative algorithms drawn from prior literature on real-time perception scheduling to contextualize performance. These baselines span a diverse spectrum of design philosophies, ranging from static heuristics to adaptive batching policies. A brief summary of each is presented below.

\begin{figure*}[ht]
    \centering
    \begin{subfigure}[t]{0.27\textwidth}
        \includegraphics[width=\linewidth]{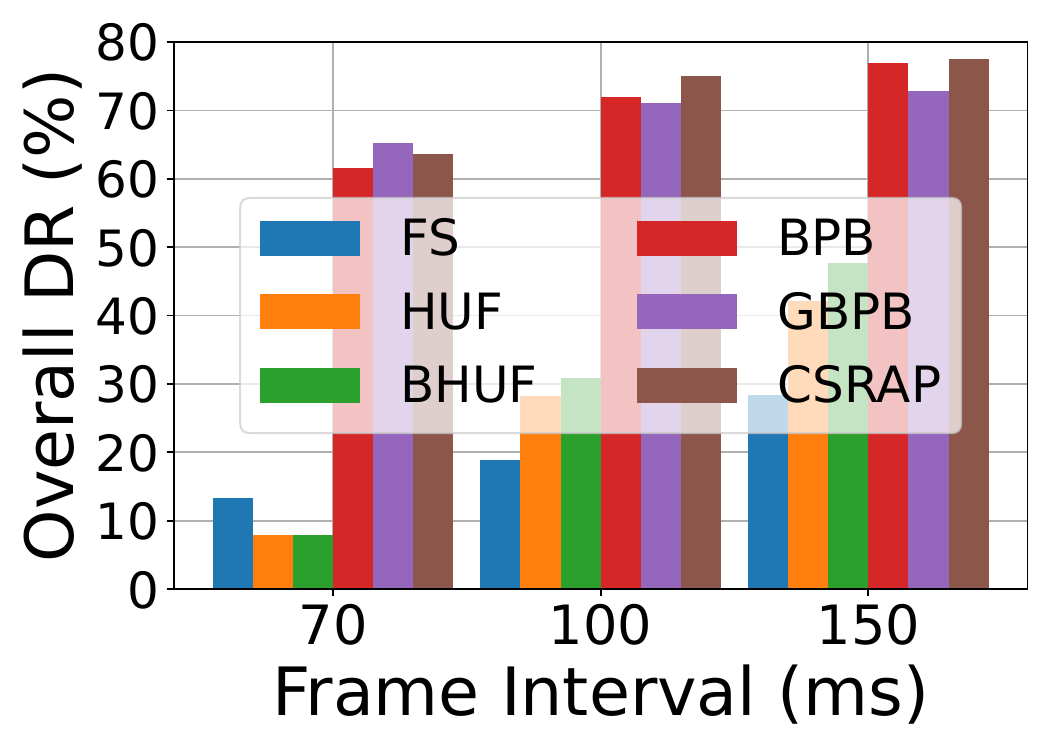}
        \caption{Overall DR.}
    \end{subfigure}
    \hfill
    \begin{subfigure}[t]{0.27\textwidth}
        \includegraphics[width=\linewidth]{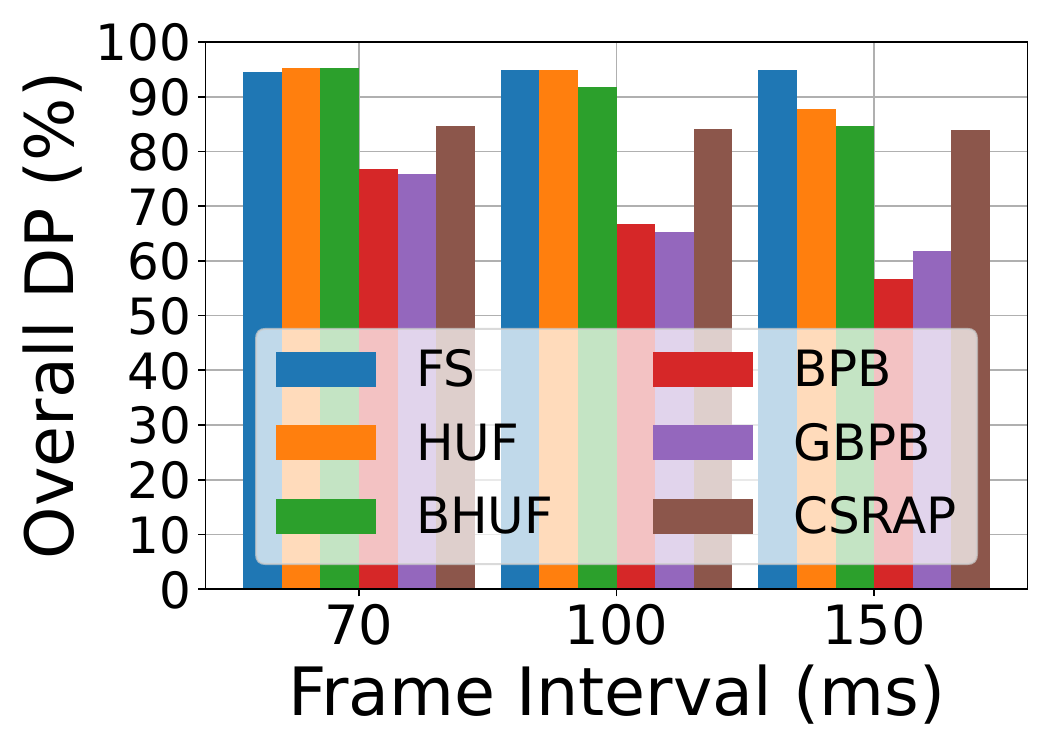}
        \caption{Overall DP.}
    \end{subfigure}
    \hfill
    \begin{subfigure}[t]{0.27\textwidth}
        \includegraphics[width=\linewidth]{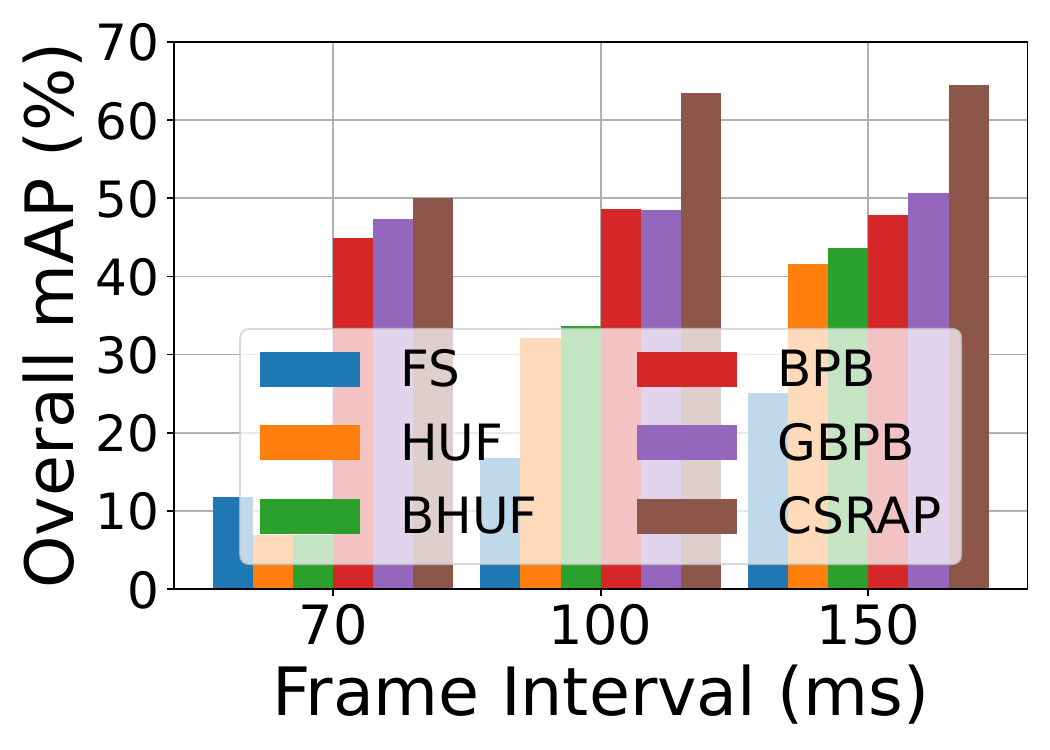}
        \caption{Overall mAP.}
    \end{subfigure}

    \begin{subfigure}[t]{0.27\textwidth}
        \includegraphics[width=\linewidth]{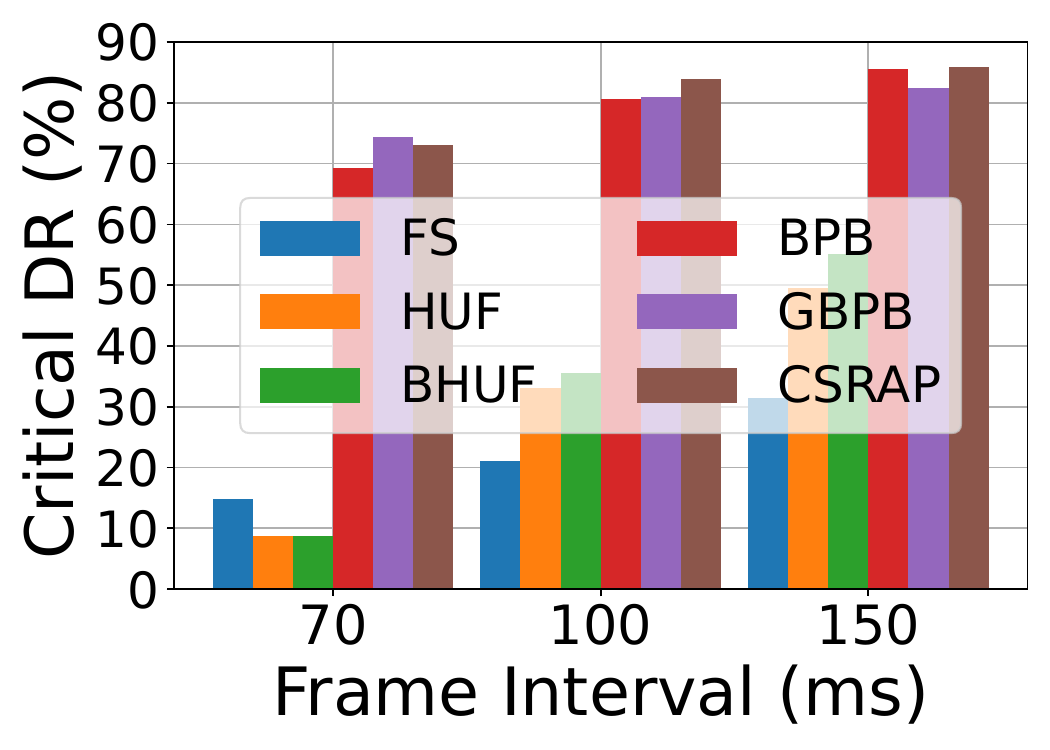}
        \caption{Critical DR.}
    \end{subfigure}
    \hfill
    \begin{subfigure}[t]{0.27\textwidth}
        \includegraphics[width=\linewidth]{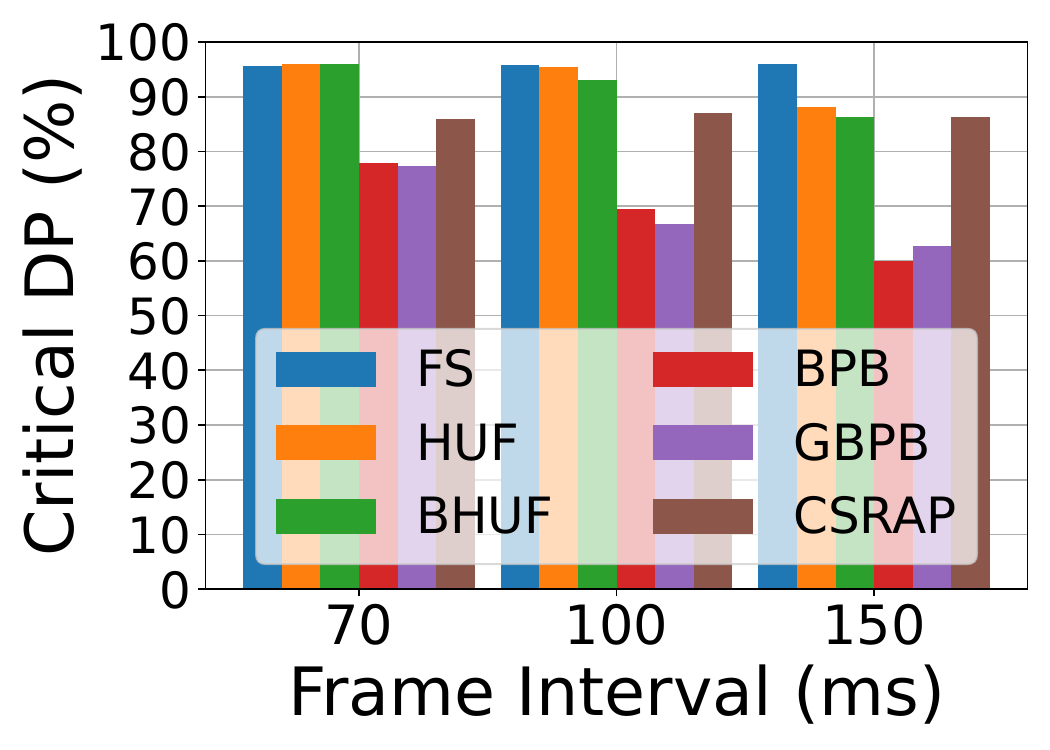}
        \caption{Critical DP.}
    \end{subfigure}
    \hfill
    \begin{subfigure}[t]{0.27\textwidth}
        \includegraphics[width=\linewidth]{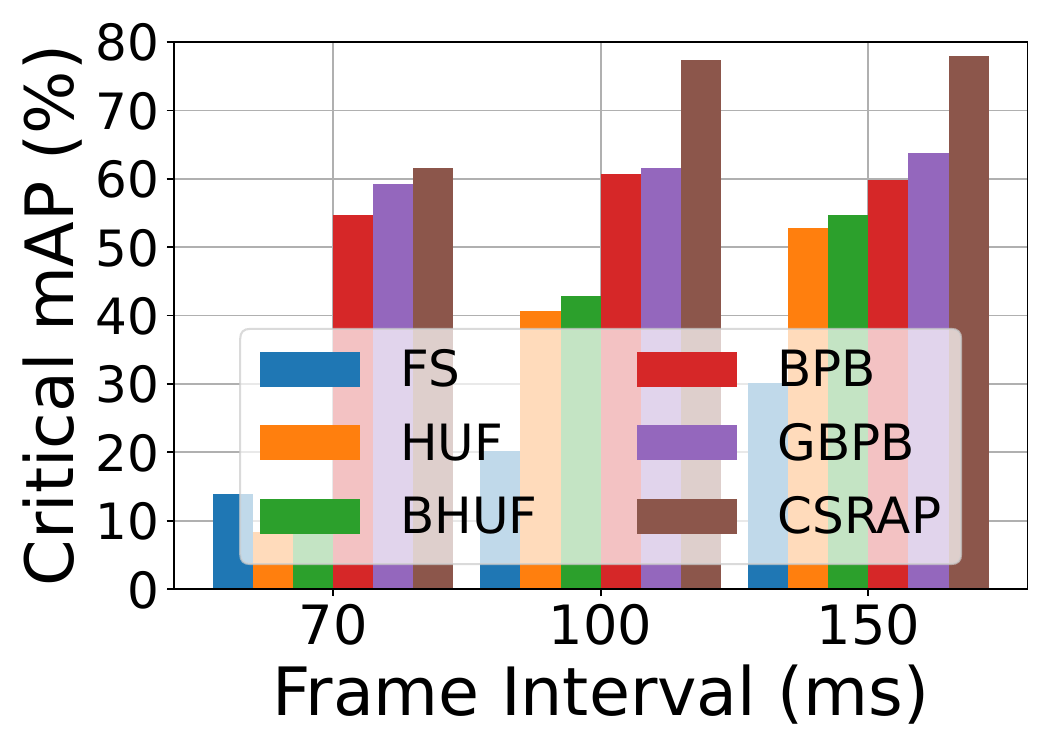}
        \caption{Critical mAP.}
    \end{subfigure}

    \caption{Comparison of scheduling algorithms on a standard traffic dataset. The first row (a to c) reports detection metrics for all objects, while the second row (d to f) highlights performance on critical objects only.}
    \label{fig:scheduling_comparison}
\end{figure*}



\begin{itemize}
\item \textbf{Full-Frame Scheduling (FS):} FS is a baseline that processes the entire frame at its original resolution. Due to the high computational cost, full-frame inference is applied only to a subset of frames, serving as a selective but high-fidelity inspection strategy.

\item \textbf{Highest Uncertainty First (HUF):} HUF greedily selects the most uncertain regions for immediate processing. It operates without batching or size-aware coordination, making decisions independently per region.

\item \textbf{Batched Highest Uncertainty First (BHUF)}~\cite{liu2020bhuf}: BHUF is an extension of HUF that incorporates batching by grouping tasks sharing the same target resolution. Task groups are constructed greedily based on uncertainty scores, enabling some computational efficiency gains.

\item \textbf{Batched Proportional Balancing (BPB)}~\cite{liu2022bpb}: BPB schedules object inspections by jointly considering uncertainty growth and class-specific criticality. It supports batching, but all inspections are performed at a fixed resolution.

\item \textbf{Generalized BPB (GBPB)}~\cite{liu2023generalized}: GBPB unifies resolution scaling, intermittent inspection, and batched execution within a single optimization framework. It extends the prior BPB scheduler by enabling fine-grained resource allocation under tight compute budgets.

\item \textbf{CSRAP:} CSRAP is the scheduling strategy introduced in Section~\ref{sec:scheduling-algo} that combines canvas-based attention scheduling with adaptive image scaling and intermittent inspection informed by object criticality.

\end{itemize}

Each baseline is executed under identical experimental conditions to ensure fairness in comparison and serves as a point of reference for quantifying the performance and efficiency gains of our proposed system.

While this work addresses key limitations of~\cite{hu2024algorithms, hu2023underprovisioned}, we do not include their method in our comparison, as their scheduler assumes a static camera and fails under camera motion—rendering it incompatible with our driving scenarios.

\begin{figure}[H]
    \centering
    \includegraphics[width=0.99\linewidth]{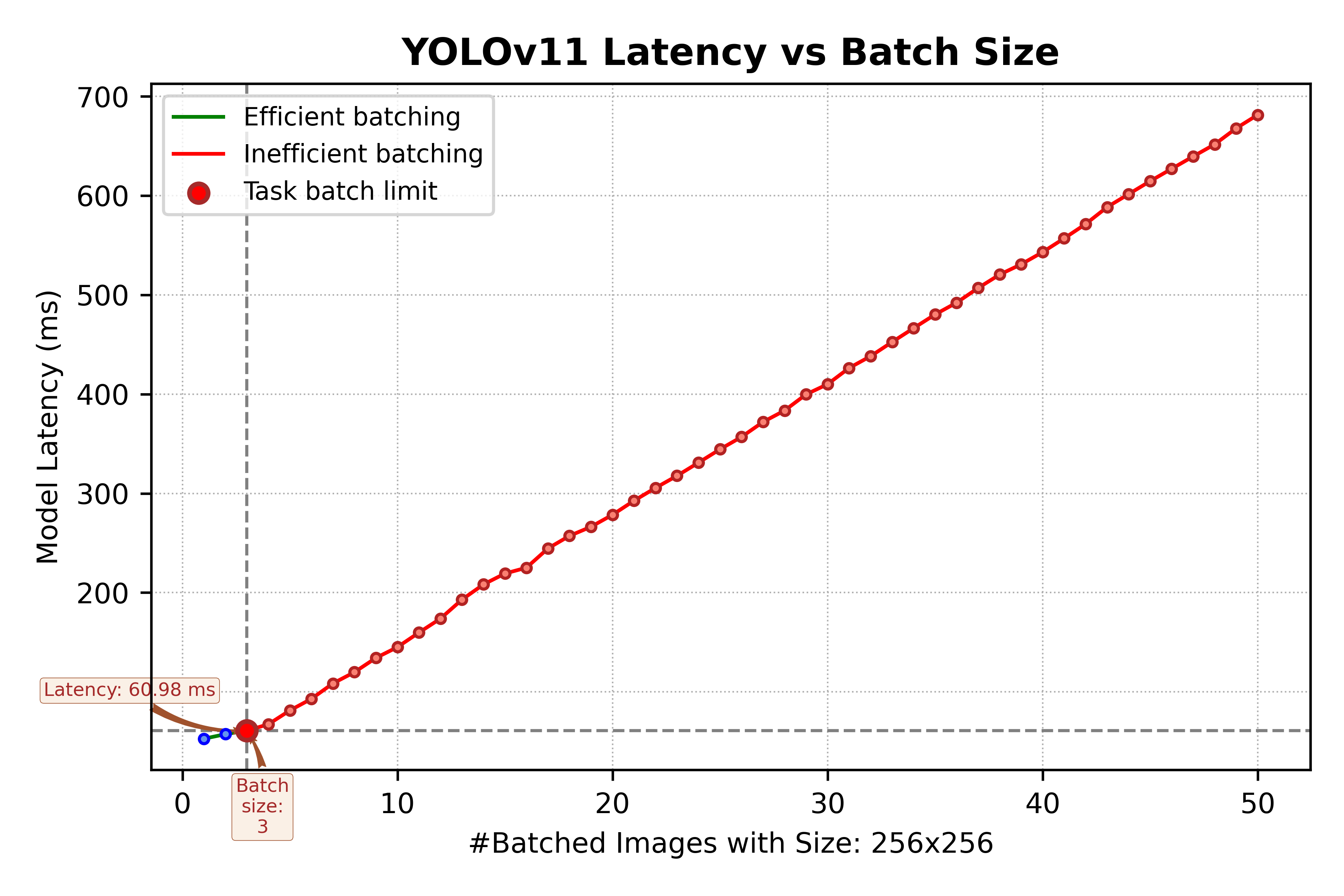}
    \caption{YOLOv11 Latency vs Batch Size for 256$\times$256 images. The graph shows model latency in milliseconds as a function of batch size. Efficient batching occurs in the lower-left region, while larger batches lead to inefficient processing and latency increases. A red marker indicates the task's optimal batch limit.}
    \label{fig:yolov11_latency}
\end{figure}

\subsection{Results}

To see how our approach holds up under different traffic loads, we took the Waymo Open Dataset and split it into two parts based on how crowded each scene is. The first slice is our "normal‑traffic" set, which features relatively sparse scenes with moderate object counts. The second slice contains high-density urban traffic, characterized by frequent occlusions and tightly clustered objects. We made sure none of those busy clips overlap with the normal set so that this "busy‑traffic" subset serves as a clean stress test, pushing the scheduler to its limits when visual clutter and decision complexity spike.

Full-frame detections on every frame—ignoring processing latency and frame period constraints—using YOLOv11-L are summarized in Table~\ref{tab:yolo11_baseline_compact} for both the normal and busy traffic datasets. These results serve as upper bounds on detection quality, providing reference points against which all adaptive scheduling methods are evaluated.

\vspace{2pt}
\begin{table}[H]
\centering
\caption{\textsc{YOLOv11L Performance on Waymo Traffic Datasets (All values in \%, latency in ms)}}
\label{tab:yolo11_baseline_compact}
\renewcommand{\arraystretch}{1.2}
\footnotesize
\resizebox{0.93\columnwidth}{!}{%
\begin{tabular}{|c|c|c||c|c|}
\hline
\multirow{2}{*}{\textbf{Metric}} & \multicolumn{2}{c||}{\textbf{Regular Traffic}} & \multicolumn{2}{c|}{\textbf{Busy Traffic}} \\
\cline{2-5}
                                 & \textbf{Overall} & \textbf{Critical} & \textbf{Overall} & \textbf{Critical} \\
\hline
Detection Recall (\%)            & 78.17   & 86.89  & 71.12   & 82.45   \\
\hline
Detection Precision (\%)         & 94.70   & 95.76   & 90.98   & 93.00   \\
\hline
mAP (\%)                         & 69.32   & 83.29   & 56.58   & 67.89   \\
\hline
Latency (ms)                     & \multicolumn{4}{c|}{419}        \\
\hline
\end{tabular}
}
\end{table}

    \begin{figure*}[ht]
    \centering
    \begin{subfigure}[t]{0.27\textwidth}
        \includegraphics[width=\linewidth]{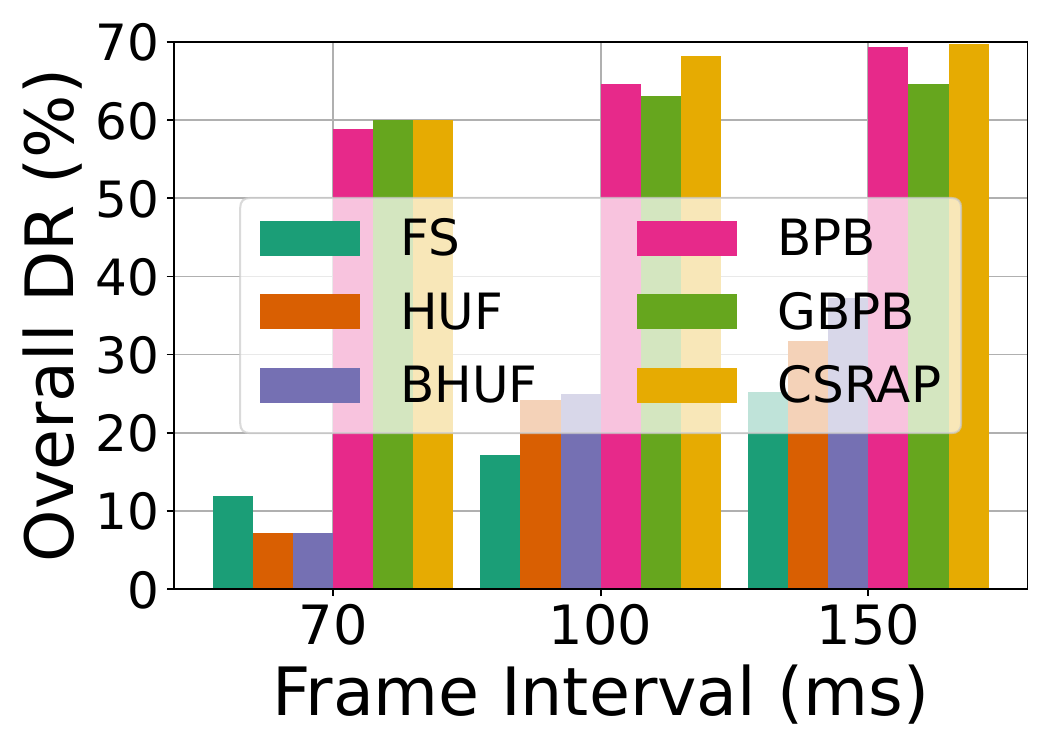}
        \caption{Overall DR.}
    \end{subfigure}
    \hfill
    \begin{subfigure}[t]{0.27\textwidth}
        \includegraphics[width=\linewidth]{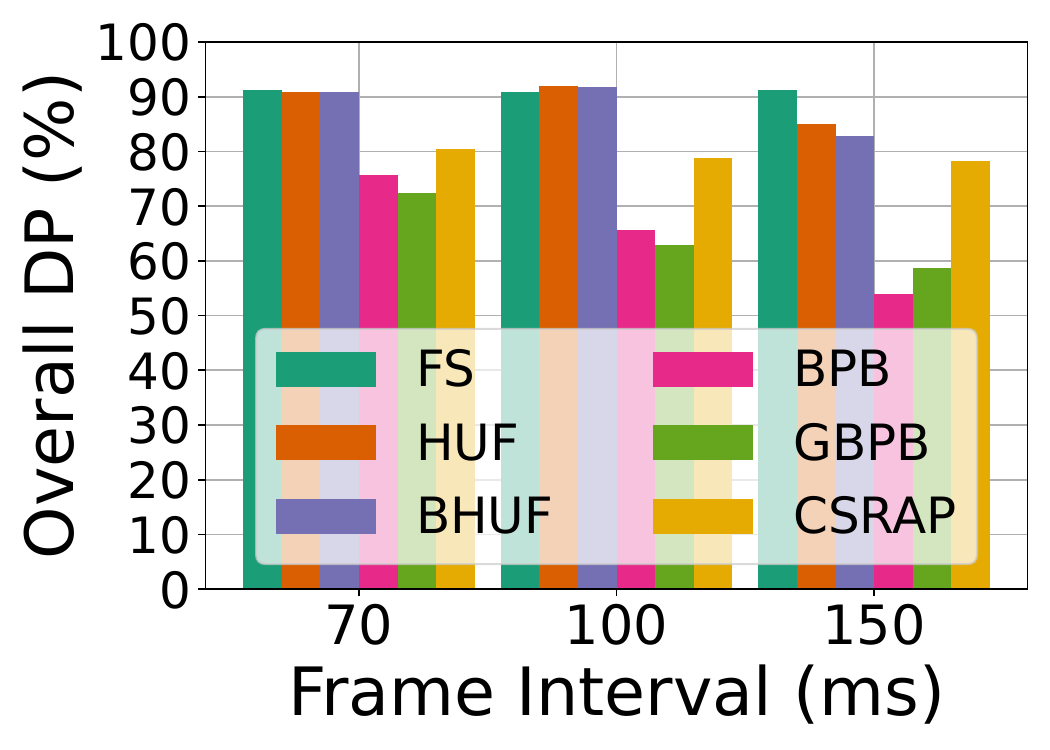}
        \caption{Overall DP.}
    \end{subfigure}
    \hfill
    \begin{subfigure}[t]{0.27\textwidth}
        \includegraphics[width=\linewidth]{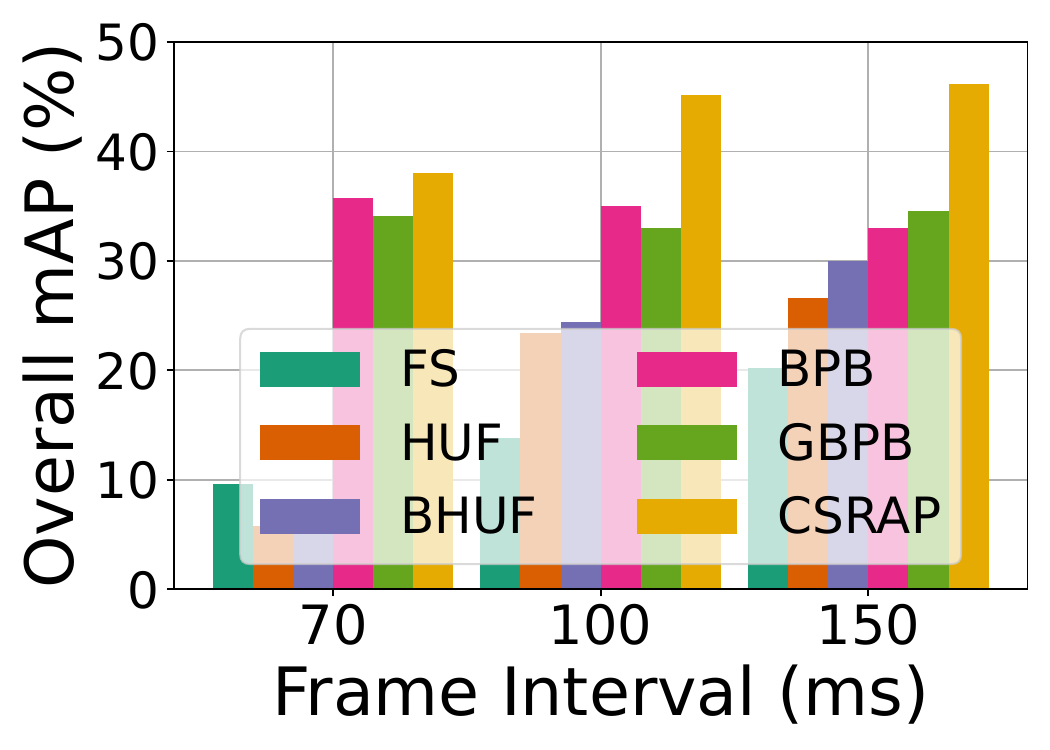}
        \caption{Overall mAP.}
    \end{subfigure}

    \begin{subfigure}[t]{0.27\textwidth}
        \includegraphics[width=\linewidth]{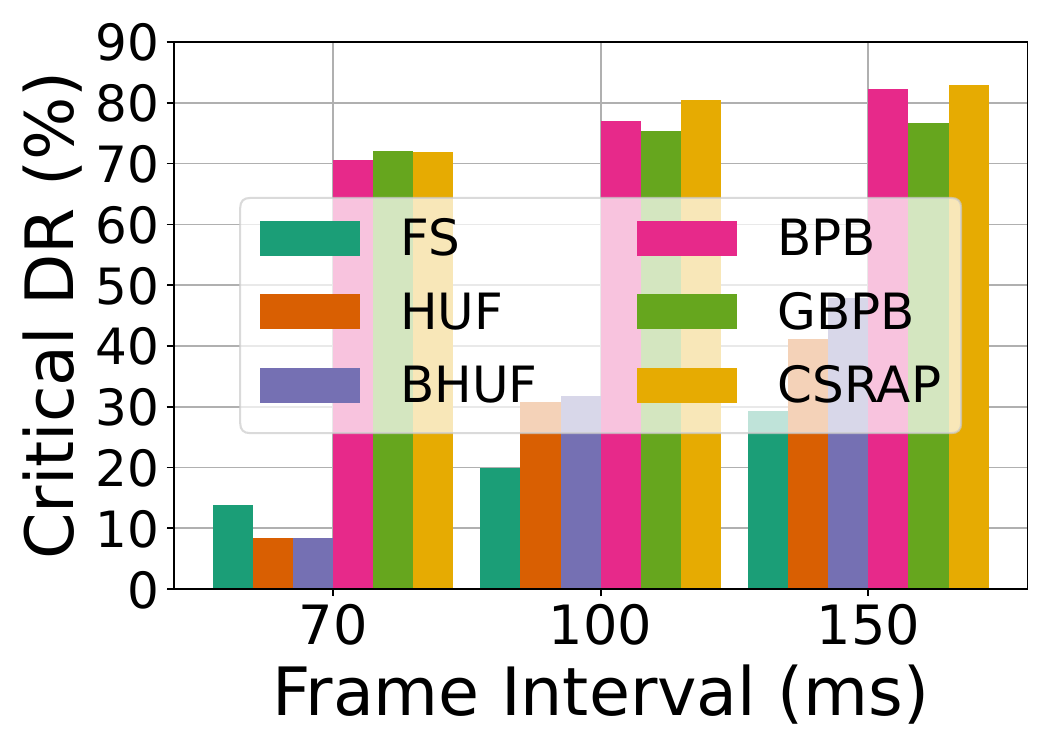}
        \caption{Critical DR.}
    \end{subfigure}
    \hfill
    \begin{subfigure}[t]{0.27\textwidth}
        \includegraphics[width=\linewidth]{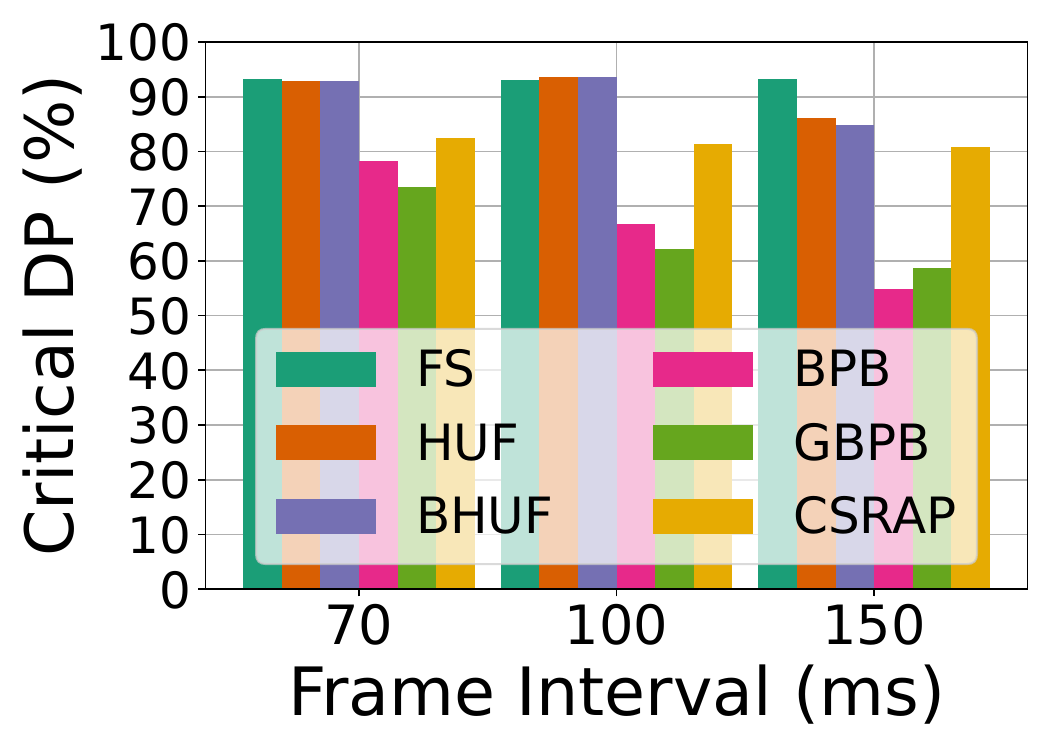}
        \caption{Critical DP.}
    \end{subfigure}
    \hfill
    \begin{subfigure}[t]{0.27\textwidth}
        \includegraphics[width=\linewidth]{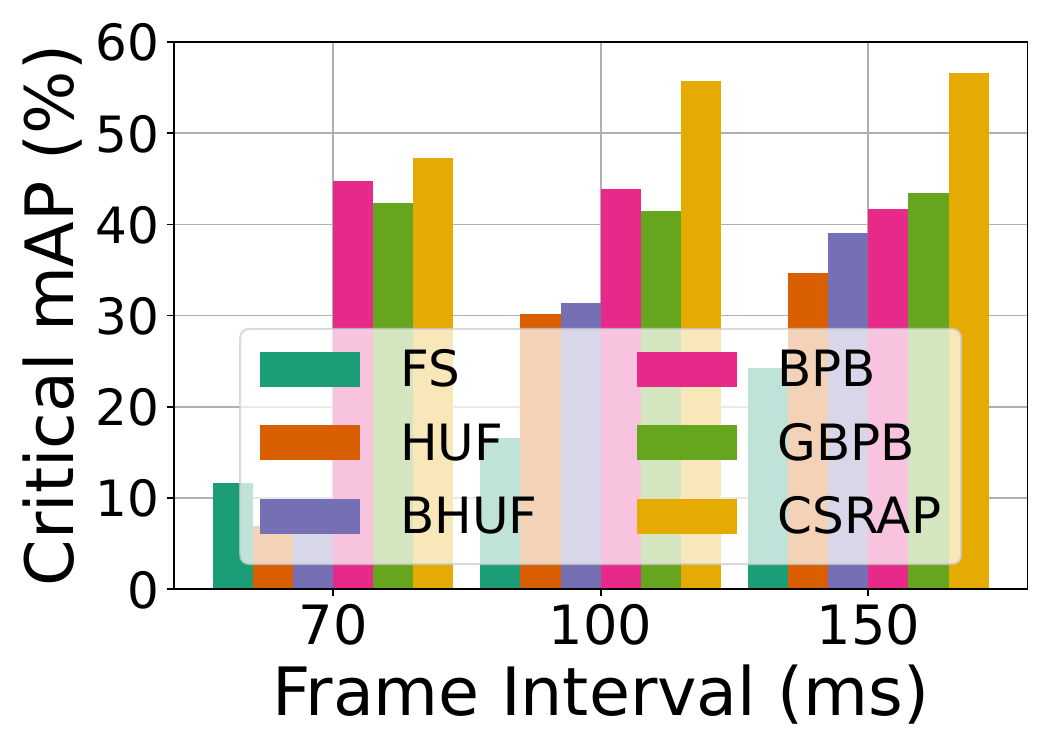}
        \caption{Critical mAP.}
    \end{subfigure}

    \caption{Comparison of scheduling algorithms in busy traffic scenarios. The first row (a to c) presents detection results for all objects, while the second row (d to f) focuses on critical objects.}
    \label{fig:busy_scheduling_comparison}
\end{figure*}



\subsubsection{Evaluation on standard traffic scenarios}
CSRAP outperforms all evaluated baselines—including GBPB, BPB, BHUF, HUF, and FS—across nearly all core metrics. The only trade-off is a slight drop in detection precision, where HUF, BHUF, and FS report marginally higher values. This is expected, as these methods detect objects on a limited set of frames without resolution scaling, preserving raw precision but at the expense of reduced recall and mAP.

The performance advantage of CSRAP stems from its architectural coherence with the realities of embedded deployment—specifically on resource-constrained systems such as the NVIDIA Jetson Orin Nano. Both PB and GPB employ task batching to enhance throughput. PB batch tasks are based on uncertainty and object importance, operating at a fixed resolution. GPB builds upon this by incorporating image downsizing and introducing resolution as an additional scheduling axis. However, both approaches implicitly assume that batching improves efficiency up to a certain point by amortizing the cost of neural inference.

On resource‑constrained edge hardware, this premise fails. As illustrated in Figure~\ref{fig:yolov11_latency}, inference latency on the Jetson Orin Nano scales nearly linearly with batch size, even for small input dimensions (e.g., $256 \times 256$). Contrary to conventional expectations, batching does not amortize inference overhead on this platform—instead, it accumulates latency across tasks, introducing substantial delays that degrade real-time responsiveness and reduce overall frame throughput.

CSRAP is explicitly designed to circumvent this limitation. Dispensing with batch processing, it adopts a canvas-based attention mechanism that schedules detection targets individually, using fine-grained control over both spatial placement and image resolution. By dynamically adjusting the resolution of each object based on its expected contribution to mission performance, CSRAP maximizes perceptual utility within a bounded inference budget. Thus, the result is higher recall on mission‑critical objects and stronger robustness under heavy computational load, which are benefits that are especially pronounced in congested traffic scenes where real‑time decisions are most demanding.

\subsubsection{Evaluation on Busy traffic scenarios}
To evaluate scheduler robustness under extreme load, we test all algorithms on the high-density traffic subset characterized by elevated object counts, occlusions, and tight spatial clustering. As expected, performance degrades across all methods due to increased computational pressure and tighter inference budgets. CSRAP, however, maintains the same performance advantage observed under standard conditions, consistently achieving superior recall and mAP. 

Batch-based schedulers like PB and GPB degrade more sharply as queuing overhead accumulates under load. While the inefficiency of batching on the Jetson Orin Nano was previously shown (Figure~\ref{fig:yolov11_latency}), its Impact is amplified in high-density scenes where delays propagate across frames. In contrast, CSRAP's canvas-based mechanism avoids batching entirely, enabling per-object control over resolution and placement. Even in the most congested frames, it sustains low-latency, high-utility inspection within timing constraints—demonstrating scalable performance under resource stress. Overall, CSRAP not only leads in standard traffic conditions but also scales more gracefully under stress, validating its design as a robust real-time scheduler for mission-critical edge perception workloads.

\subsubsection{Evaluation on Physically Close Objects}

\vspace{-9pt}
\begin{figure}[htbp]
  \centering
  \includegraphics[width=0.99\linewidth]{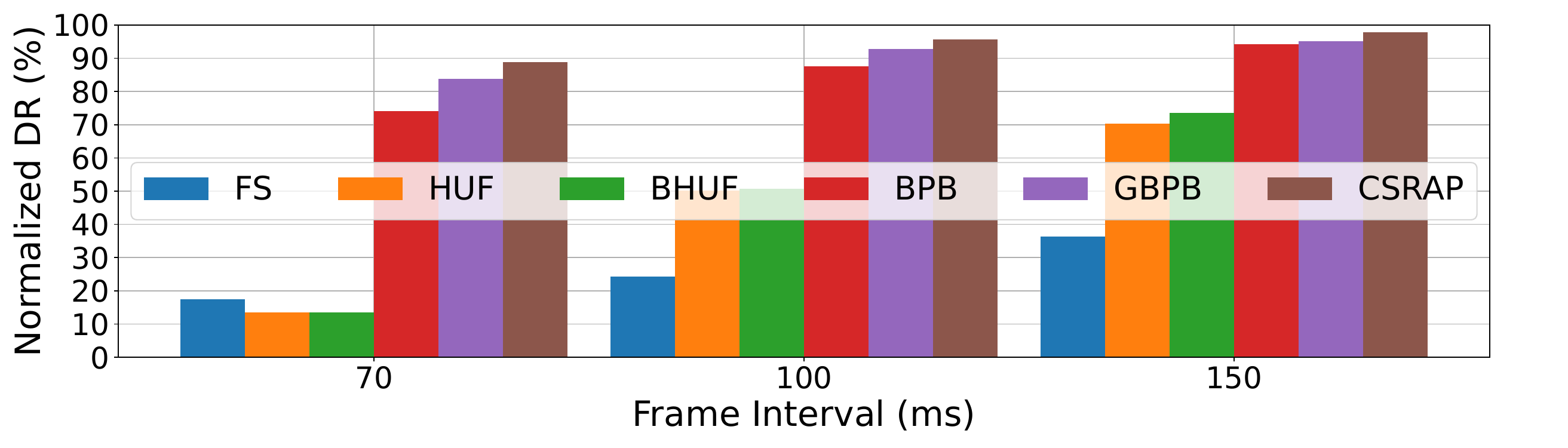}
  \caption{Normalized Detection Recall (DR) for close objects across schedulers}
  \label{fig:response_to_close_objects}
\end{figure}

\vspace{-9pt}
\begin{figure}[!htbp]  
  \centering
  \begin{subfigure}[t]{0.328\linewidth}
    \centering
    \includegraphics[width=\linewidth]{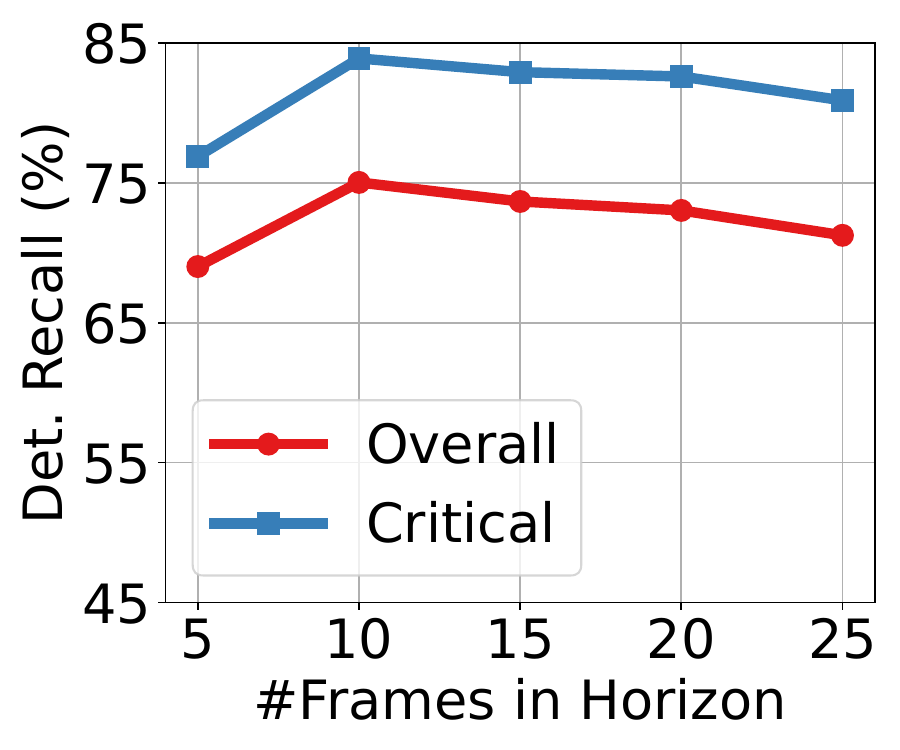}
    \caption{Detection recall}
    \label{fig:csrap_horizon_recall}
  \end{subfigure}\hfill
  \begin{subfigure}[t]{0.328\linewidth}
    \centering
    \includegraphics[width=\linewidth]{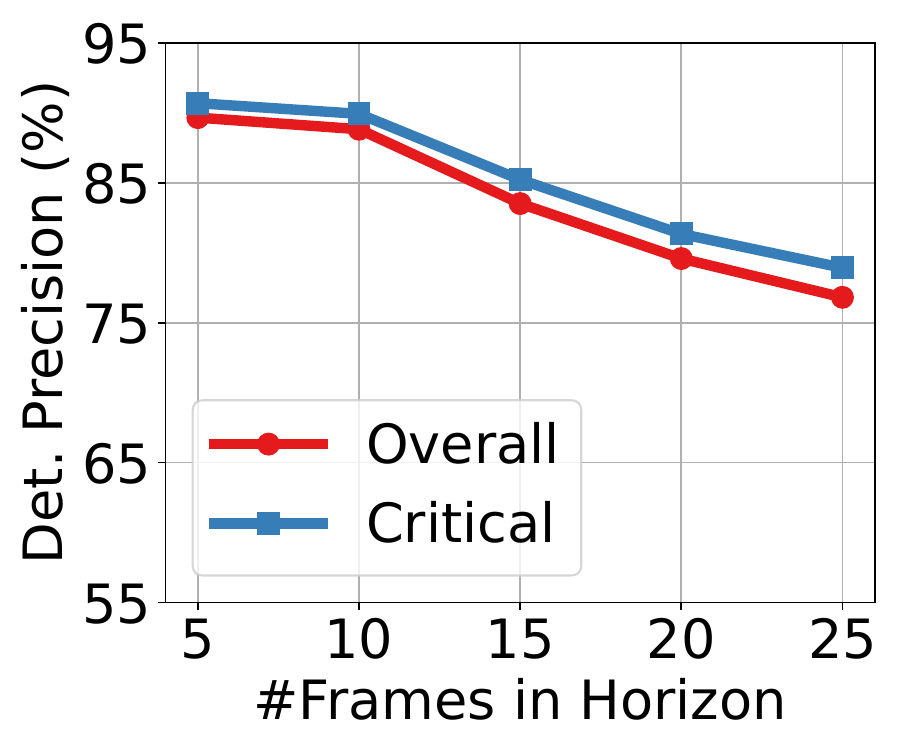}
    \caption{Detection precision}
    \label{fig:csrap_horizon_precision}
  \end{subfigure}\hfill
  \begin{subfigure}[t]{0.328\linewidth}
    \centering
    \includegraphics[width=\linewidth]{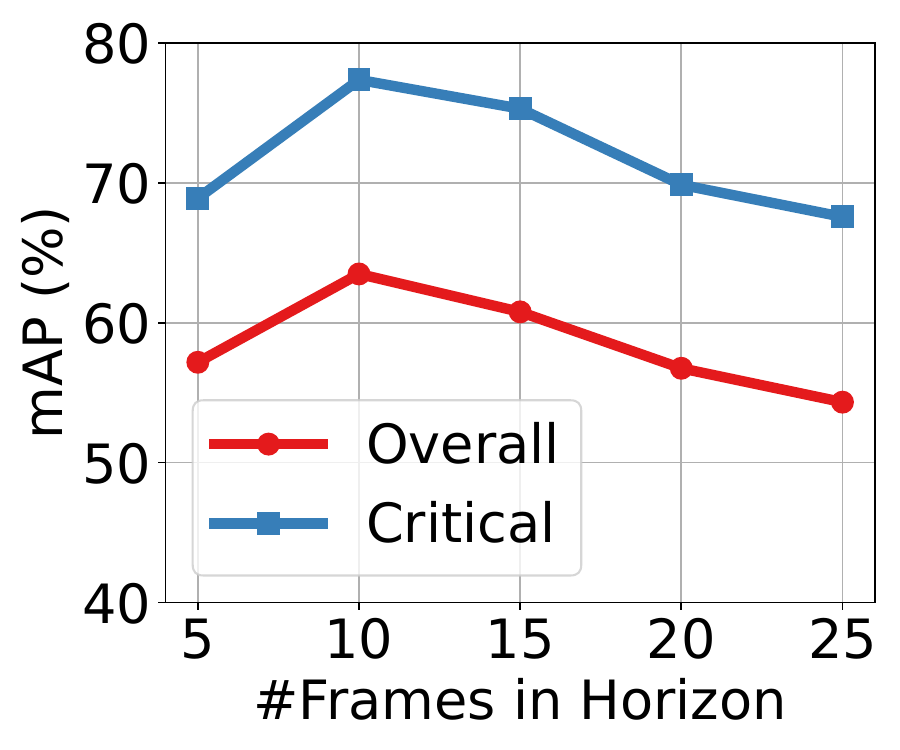}
    \caption{mAP}
    \label{fig:csrap_horizon_map}
  \end{subfigure}

  \caption{Impact of horizon length on CSRAP detection accuracy metrics for overall and critical objects.}
  \label{fig:csrap_horizon_metrics}
\end{figure}

To assess perception fidelity in the near field, we measure detection recall for objects within 30 meters of the ego vehicle, normalized against a full-frame inspection on every frame (Figure~\ref{fig:response_to_close_objects}). CSRAP consistently achieves the highest normalized recall across all frame intervals, with the margin most pronounced under the tightest timing constraint (70 ms). This demonstrates CSRAP's ability to reliably prioritize and inspect close-range, mission-critical objects even when scheduling flexibility is minimal. As frame intervals increase, baseline schedulers show moderate improvements due to reduced compute contention, yet CSRAP remains superior. The narrowing margin reflects early saturation in CSRAP's performance, not degradation—underscoring its robustness under stringent real-time constraints.

These results affirm CSRAP's core design principle: leveraging visual scale as a proxy for proximity to guide attention without requiring depth sensors. By dynamically allocating resolution based on both object size and criticality, CSRAP sustains high recall on near-field targets. This spatial adaptivity gives it a decisive advantage over global resizing or batching-based methods that lack fine-grained prioritization.

\vspace{-1pt}
\subsubsection{Evaluation using Varying Scheduling Horizon Length}

To assess how horizon length impacts CSRAP's effectiveness, we vary it from 5 to 25 frames while fixing the frame interval at 0.1 seconds. As shown in Figure~\ref{fig:csrap_horizon_metrics}, CSRAP maintains stable detection performance across this range, demonstrating strong robustness. Peak recall occurs near a 10-frame horizon. Shorter horizons increase full-frame inspection frequency, reducing partial-frame flexibility. Longer horizons delay re-evaluation, risking outdated detections and inefficient compute utilization. This trade-off reflects a balance between temporal freshness and scheduling adaptability. A 10-frame horizon offers the best compromise and is adopted as the system default.

\begin{figure}[htbp]
  \centering
  \begin{subfigure}[b]{0.49\linewidth}
    \centering
    \includegraphics[width=\linewidth]{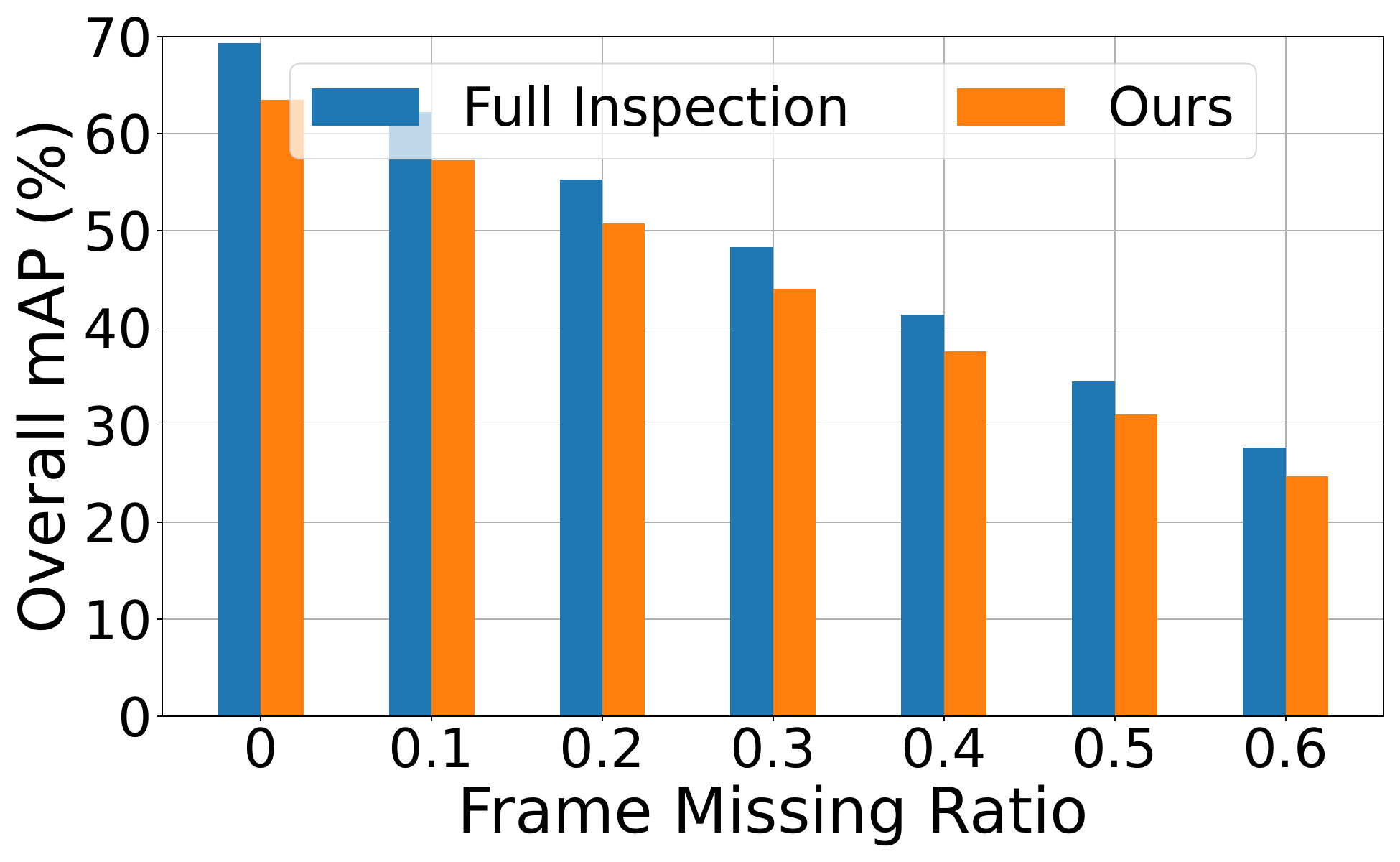}
    \caption{Overall mAP under varying frame missing ratios.}
    \label{fig:fmr_overall}
  \end{subfigure}
  \hfill
  \begin{subfigure}[b]{0.49\linewidth}
    \centering
    \includegraphics[width=\linewidth]{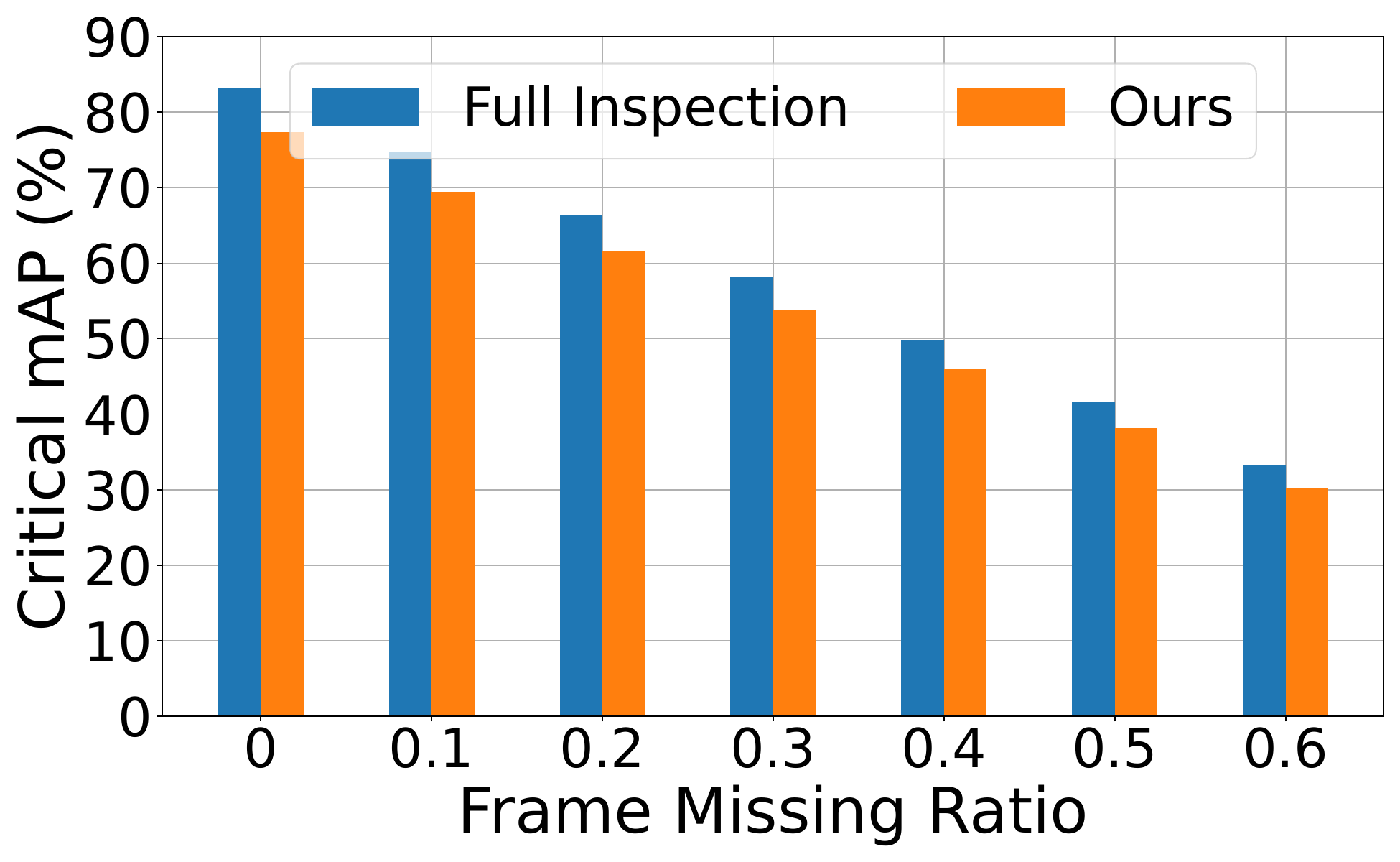}
    \caption{Critical object mAP under varying frame missing ratios.}
    \label{fig:fmr_critical}
  \end{subfigure}
  \caption{Comparison of detection accuracy for Full Inspection vs.\ CSRAP under increasing frame missing ratios.}
  \label{fig:fmr_comparison}
\end{figure}

\subsubsection{Impact of Missing Frames}

To evaluate CSRAP's resilience under data degradation, we simulate frame loss by randomly dropping 10–60\% of frames within each 10-frame scheduling horizon, applying identical deletion patterns to both CSRAP and the full-frame baseline using a fixed random seed. If a keyframe is missing, CSRAP automatically promotes the next available frame as the scheduling anchor. As shown in Figure~\ref{fig:fmr_comparison}, CSRAP sustains both overall and critical-object mAP closely aligned with the baseline, even under 60\% frame loss. This robustness stems from two core mechanisms, which are optical flow interpolation between non-consecutive frames and conservative candidate region expansion to mitigate temporal uncertainty. These results confirm the ability of CSRAP to preserve detection fidelity under severe data sparsity, making it a strong fit for mission-critical deployments subject to bandwidth constraints, transmission failures, or intermittent sensing.

\begin{figure}[htbp]
  \centering

    \begin{subfigure}[b]{0.87\linewidth}
    \centering
    \includegraphics[width=\linewidth]{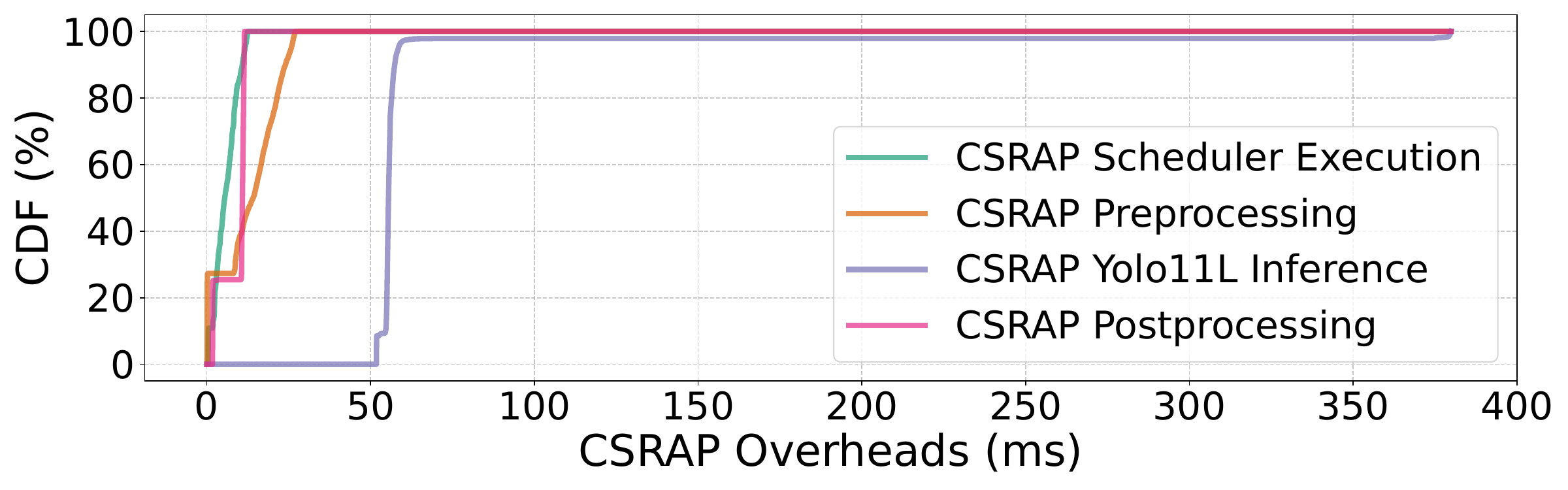}
    \caption{CSRAP per-stage latency CDF.}
    \label{fig:cdf_csrap_metrics}
  \end{subfigure}

  \begin{subfigure}[b]{0.87\linewidth}
    \centering
    \includegraphics[width=\linewidth]{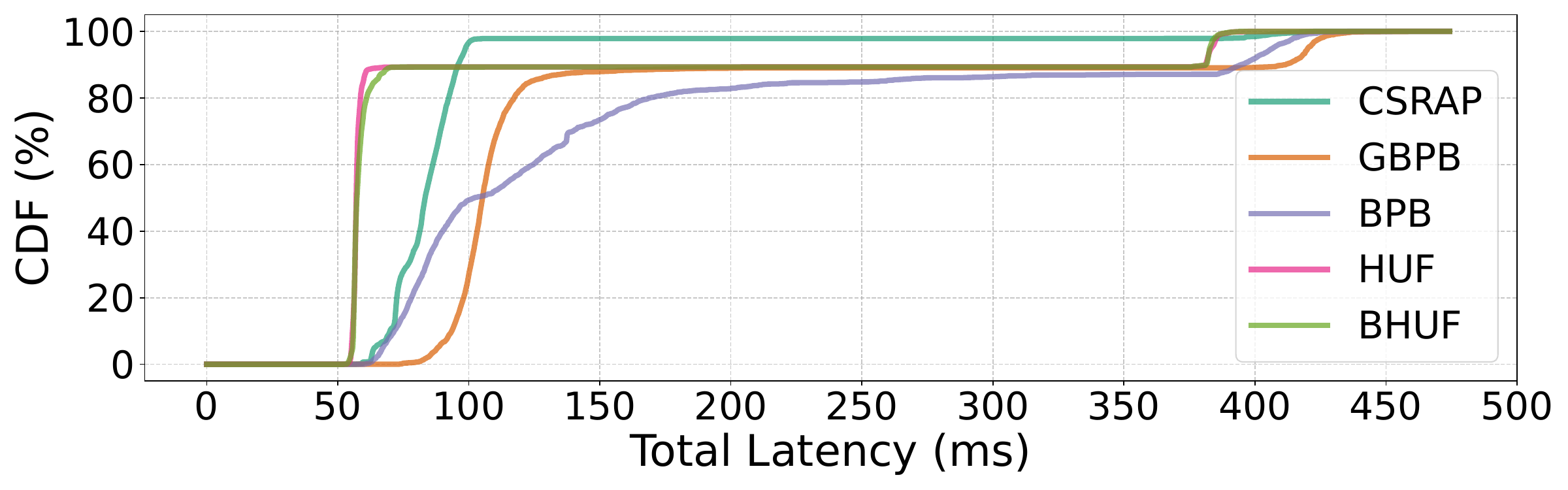}
    \caption{End-to-end latency CDF across schedulers.}
    \label{fig:cdf_total_delay}
  \end{subfigure}

  \caption{Latency evaluation of scheduling algorithms and their components. (a) CSRAP's internal overheads per pipeline stage. (b) Total delay distributions across schedulers. }
  \label{fig:scheduler_latency_breakdown}
\end{figure}

\subsubsection{Breakdown of Overheads}
We profile the end-to-end per-frame latency of each scheduler, as shown in Figure~\ref{fig:scheduler_latency_breakdown}, and observe that CSRAP achieves the lowest overall delay at 78.3 ms—outperforming PB (104.3 ms) and GPB (107.5 ms), and nearing the efficiency of greedy and serialized baselines. This is enabled by CSRAP's lightweight pipeline in which global and frame-level scheduling requires only 0.58 ms and 11.3 ms, respectively, while inference latency remains low at 59.1 ms due to resolution-aware canvas construction and selective downscaling. Although postprocessing adds 7.3 ms from coordinate remapping, this overhead is minimal relative to the 100–150 ms frame interval. In comparison, PB and GPB experience batching-induced latency growth due to non-linear scaling on edge hardware. By avoiding batching altogether, CSRAP maintains real-time responsiveness across load conditions, validating its suitability for latency-critical edge perception.
\vspace{-9pt}
\section{Related Work}
\label{sec:related_work}
\vspace{-2pt}
Real-time perception in mission-critical systems has driven a shift toward adaptive scheduling mechanisms that reconcile deep learning with stringent system-level constraints. Autonomous platforms now routinely embed deep neural networks (DNNs) within feedback loops~\cite{KiranSTMSYP22, HuangCYW24, UppuluriPMKC25, LuMSY25, NanZYZZD25}, where latency and resource efficiency are as vital as inference accuracy. Yet conventional DNNs are designed for best-effort throughput, lacking intrinsic support for bounded execution time, resource-aware reconfiguration, or deterministic behavior—rendering them inadequate for cyber-physical systems (CPS) with hard real-time guarantees.

To address these challenges, recent work has explored real-time scheduling at multiple layers of the perception stack. System-level schedulers seek to optimize CPU–GPU coordination and pipeline efficiency through task partitioning and compute stage orchestration~\cite{CittadiniMB25, AmertOYAS17, 3724420, CapodieciCBP18, OlmedoCMMB20}. At the model level, techniques compress~\cite{YaoZSLLSA18, ZhuZL18, Yao0LWLSA20, ZhangWXD24, LiCLZTL25, DaiFMH25} or selectively reconfigure~\cite{DBLP:conf/rtss/Bateni018, DBLP:conf/rtas/LeeN20, 3724126, YaoHZSLLW0A20, XuLFL0C24, SoyyigitYY24} DNN architectures to maintain performance under varying latency constraint. 

In recent years, data-centric scheduling has emerged as a powerful third axis of optimization, improving efficiency by concentrating computation on spatial regions that are semantically relevant. Attention scheduling~\cite{liu2020removing} tackles the problem of priority inversion by dynamically reordering which parts of the frame are processed first. However, many existing approaches ~\cite{liu2020removing, LiuYFTYBYSA2022, HuLAWD21} rely on auxiliary sensors such as LiDAR to estimate saliency. This dependence complicates system design and limits practical deployment, especially on resource-constrained edge devices. Broadly speaking, existing work spans three main scheduling strategies—spatial~\cite{LiuYFTYBYSA2020, LiuYFTYBYSA2022}, temporal~\cite{HuLAWD21, YaoZZSA17, YaoHZSLLW0A20, YaoZSLLSA18}, and quality-adaptive~\cite{teerapittayanon2016branchynet}—to strike a balance between accuracy and resource efficiency. Canvas-based methods~\cite{hu2023underprovisioned, hu2024algorithms, Ila_Gokarn_2024_1, Ila_Gokarn_2024_2} optimize computation by packing quantized object regions into fixed-size canvases for EDF-based inference. While this improves throughput, these methods rigidly assume fixed viewpoints and overlook temporal object location uncertainty. On the other hand, uncertainty-aware strategies~\cite{liu2023generalized, liu2022self} adjust how often objects are inspected based on predicted motion, but they still operate on uniformly resized and quantized batches, which reduces their flexibility in handling scenes with high spatial or temporal variability.

CSRAP addresses these limitations through a unified, self-cueing framework that integrates adaptive object resizing, spatiotemporal attention fusion, and internal uncertainty estimation—without requiring external saliency signals. Its support for multi-resolution canvas scheduling and dynamic inspection frequency generalizes prior approaches, offering scalable, low-latency perception under constrained compute budgets.
\vspace{-6pt}
\section{Conclusion}
\label{sec:conclusion}
\vspace{-3pt}
The paper presented an extension to canvas-based scheduling that allows a more flexible selection of canvas frame sizes and rates compared to previous solutions. The new algorithm was compared to prior canvas-based scheduling algorithms as well as to state-of-the-art batching baselines, demonstrating an improvement in mAP and recall across a range of evaluated scenarios. The work demonstrates the importance of investigating innovative solutions for selective filtering of input data (to reduce load) in edge AI applications and the need for fitting selected data objects to appropriately chosen templates (canvas frames) that can be efficiently handled by lower-end GPUs. Future work will explore multimodal and distributed extensions to the problem, where multiple data streams of differing characteristics must be jointly processed.
\section*{Acknowledgment}

Generative AI was used exclusively for language refinement and grammatical checks. All technical content, including results, data analysis, and code, was developed independently by the authors.

\bibliographystyle{ieeetr}
\bibliography{Main}

\begin{thebibliography}{10}

\bibitem{ren2015towards}
S.~Ren, K.~He, R.~Girshick, J.~Sun, and R.~Faster, ``Towards real-time object detection with region proposal networks, adv,'' {\em Neural Inf. Process}, vol.~28, 2015.

\bibitem{redmon2016you}
J.~Redmon, S.~Divvala, R.~Girshick, and A.~Farhadi, ``You only look once: Unified, real-time object detection,'' in {\em Proceedings of the IEEE conference on computer vision and pattern recognition}, pp.~779--788, 2016.

\bibitem{liu2016ssd}
W.~Liu, D.~Anguelov, D.~Erhan, C.~Szegedy, S.~E. Reed, C.~Fu, and A.~C. Berg, ``{SSD:} single shot multibox detector,'' in {\em Computer Vision - {ECCV} 2016 - 14th European Conference, Amsterdam, The Netherlands, October 11-14, 2016, Proceedings, Part {I}} (B.~Leibe, J.~Matas, N.~Sebe, and M.~Welling, eds.), vol.~9905 of {\em Lecture Notes in Computer Science}, pp.~21--37, Springer, 2016.

\bibitem{liu2020removing}
S.~Liu, S.~Yao, X.~Fu, R.~Tabish, S.~Yu, A.~Bansal, H.~Yun, L.~Sha, and T.~Abdelzaher, ``On removing algorithmic priority inversion from mission-critical machine inference pipelines,'' in {\em 2020 IEEE Real-Time Systems Symposium (RTSS)}, pp.~319--332, IEEE, 2020.

\bibitem{liu2022self}
S.~Liu, X.~Fu, M.~Wigness, P.~David, S.~Yao, L.~Sha, and T.~Abdelzaher, ``Self-cueing real-time attention scheduling in criticality-aware visual machine perception,'' in {\em 2022 IEEE 28th Real-Time and Embedded Technology and Applications Symposium (RTAS)}, pp.~173--186, IEEE, 2022.

\bibitem{hu2023underprovisioned}
Y.~Hu, I.~Gokarn, S.~Liu, A.~Misra, and T.~F. Abdelzaher, ``Underprovisioned gpus: On sufficient capacity for real-time mission-critical perception,'' in {\em 32nd International Conference on Computer Communications and Networks, {ICCCN} 2023, Honolulu, HI, USA, July 24-27, 2023}, pp.~1--10, {IEEE}, 2023.

\bibitem{hu2024algorithms}
Y.~Hu, I.~Gokarn, S.~Liu, A.~Misra, and T.~F. Abdelzaher, ``Algorithms for canvas-based attention scheduling with resizing,'' in {\em 30th {IEEE} Real-Time and Embedded Technology and Applications Symposium, {RTAS} 2024, Hong Kong, May 13-16, 2024}, pp.~348--359, {IEEE}, 2024.

\bibitem{liu2023generalized}
S.~Liu, X.~Fu, Y.~Hu, M.~B. Wigness, P.~David, S.~Yao, L.~Sha, and T.~F. Abdelzaher, ``Generalized self-cueing real-time attention scheduling with intermittent inspection and image resizing,'' {\em Real Time Syst.}, vol.~59, no.~2, pp.~302--343, 2023.

\bibitem{liu2022bpb}
S.~Liu, X.~Fu, M.~B. Wigness, P.~David, S.~Yao, L.~Sha, and T.~F. Abdelzaher, ``Self-cueing real-time attention scheduling in criticality-aware visual machine perception,'' in {\em 28th {IEEE} Real-Time and Embedded Technology and Applications Symposium, {RTAS} 2022, Milano, Italy, May 4-6, 2022}, pp.~173--186, {IEEE}, 2022.

\bibitem{KroegerTDG16}
T.~Kroeger, R.~Timofte, D.~Dai, and L.~V. Gool, ``Fast optical flow using dense inverse search,'' in {\em Computer Vision - {ECCV} 2016 - 14th European Conference, Amsterdam, The Netherlands, October 11-14, 2016, Proceedings, Part {IV}} (B.~Leibe, J.~Matas, N.~Sebe, and M.~Welling, eds.), vol.~9908 of {\em Lecture Notes in Computer Science}, pp.~471--488, Springer, 2016.

\bibitem{kt1}
H.~W. Kuhn and A.~W. Tucker, {\em Nonlinear Programming}.
\newblock Berkeley, California: University of California Press, 1951.

\bibitem{CoffmanCGJSWY00}
E.~G.~C. Jr., C.~Courcoubetis, M.~R. Garey, D.~S. Johnson, P.~W. Shor, R.~R. Weber, and M.~Yannakakis, ``Bin packing with discrete item sizes, {Part I}: Perfect packing theorems and the average case behavior of optimal packings,'' {\em SIAM Journal on Discrete Mathematics}, vol.~13, no.~3, pp.~384--402, 2000.

\bibitem{pr1}
J.~Januszewski, ``Packing rectangles into the unit square,'' {\em Geometriae Dedicata}, vol.~81, pp.~13--18, 01 2000.

\bibitem{waymoR}
P.~Sun, H.~Kretzschmar, X.~Dotiwalla, A.~Chouard, V.~Patnaik, P.~Tsui, J.~Guo, Y.~Zhou, Y.~Chai, B.~Caine, V.~Vasudevan, W.~Han, J.~Ngiam, H.~Zhao, A.~Timofeev, S.~Ettinger, M.~Krivokon, A.~Gao, A.~Joshi, Y.~Zhang, J.~Shlens, Z.~Chen, and D.~Anguelov, ``Scalability in perception for autonomous driving: Waymo open dataset,'' in {\em 2020 IEEE/CVF Conference on Computer Vision and Pattern Recognition (CVPR)}, pp.~2443--2451, 2020.

\bibitem{yoloR1}
J.~Redmon, S.~K. Divvala, R.~B. Girshick, and A.~Farhadi, ``You only look once: Unified, real-time object detection,'' in {\em 2016 {IEEE} Conference on Computer Vision and Pattern Recognition, {CVPR} 2016, Las Vegas, NV, USA, June 27-30, 2016}, pp.~779--788, {IEEE} Computer Society, 2016.

\bibitem{yoloR2}
G.~Jocher, J.~Qiu, and A.~Chaurasia, ``{Ultralytics YOLO},'' Jan. 2023.

\bibitem{cocoR}
T.-Y. Lin, M.~Maire, S.~Belongie, J.~Hays, P.~Perona, D.~Ramanan, P.~Doll{\'a}r, and C.~L. Zitnick, ``Microsoft coco: Common objects in context,'' in {\em Computer Vision -- ECCV 2014} (D.~Fleet, T.~Pajdla, B.~Schiele, and T.~Tuytelaars, eds.), (Cham), pp.~740--755, Springer International Publishing, 2014.

\bibitem{mAPR}
J.~{Cartucho}, R.~{Ventura}, and M.~{Veloso}, ``Robust object recognition through symbiotic deep learning in mobile robots,'' in {\em 2018 IEEE/RSJ International Conference on Intelligent Robots and Systems (IROS)}, pp.~2336--2341, 2018.

\bibitem{liu2020bhuf}
L.~Liu, Z.~Dong, Y.~Wang, and W.~Shi, ``Prophet: Realizing a predictable real-time perception pipeline for autonomous vehicles,'' in {\em {IEEE} Real-Time Systems Symposium, {RTSS} 2022, Houston, TX, USA, December 5-8, 2022}, pp.~305--317, {IEEE}, 2022.

\bibitem{KiranSTMSYP22}
B.~R. Kiran, I.~Sobh, V.~Talpaert, P.~Mannion, A.~A.~A. Sallab, S.~K. Yogamani, and P.~P{\'{e}}rez, ``Deep reinforcement learning for autonomous driving: {A} survey,'' {\em {IEEE} Trans. Intell. Transp. Syst.}, vol.~23, no.~6, pp.~4909--4926, 2022.

\bibitem{HuangCYW24}
X.~Huang, Y.~Cheng, Q.~Yu, and X.~Wang, ``Deep reinforcement learning for autonomous driving based on safety experience replay,'' {\em {IEEE} Trans. Cogn. Dev. Syst.}, vol.~16, no.~6, pp.~2070--2084, 2024.

\bibitem{UppuluriPMKC25}
B.~Uppuluri, A.~Patel, N.~Mehta, S.~Kamath, and P.~Chakraborty, ``Curla: Curriculum learning based deep reinforcement learning for autonomous driving,'' in {\em Proceedings of the 17th International Conference on Agents and Artificial Intelligence, {ICAART} 2025 - Volume 3, Porto, Portugal, February 23-25, 2025} (A.~P. Rocha, L.~Steels, and H.~J. van~den Herik, eds.), pp.~435--442, {SCITEPRESS}, 2025.

\bibitem{LuMSY25}
Y.~Lu, H.~Ma, E.~Smart, and H.~Yu, ``Enhancing autonomous driving decision: {A} hybrid deep reinforcement learning-kinematic-based autopilot framework for complex motorway scenes,'' {\em {IEEE} Trans. Intell. Transp. Syst.}, vol.~26, no.~3, pp.~3198--3209, 2025.

\bibitem{NanZYZZD25}
J.~Nan, R.~Zhang, G.~Yin, W.~Zhuang, Y.~Zhang, and W.~Deng, ``Safe and interpretable human-like planning with transformer-based deep inverse reinforcement learning for autonomous driving,'' {\em {IEEE} Trans Autom. Sci. Eng.}, vol.~22, pp.~12134--12146, 2025.

\bibitem{CittadiniMB25}
E.~Cittadini, M.~Marinoni, and G.~C. Buttazzo, ``A hardware accelerator to support deep learning processor units in real-time image processing,'' {\em Eng. Appl. Artif. Intell.}, vol.~145, p.~110159, 2025.

\bibitem{AmertOYAS17}
T.~Amert, N.~Otterness, M.~Yang, J.~H. Anderson, and F.~D. Smith, ``{GPU} scheduling on the {NVIDIA} {TX2:} hidden details revealed,'' in {\em 2017 {IEEE} Real-Time Systems Symposium, {RTSS} 2017, Paris, France, December 5-8, 2017}, pp.~104--115, {IEEE} Computer Society, 2017.

\bibitem{3724420}
X.~Wang, Z.~Tang, J.~Guo, T.~Meng, C.~Wang, T.~Wang, and W.~Jia, ``Empowering edge intelligence: A comprehensive survey on on-device ai models,'' {\em ACM Comput. Surv.}, vol.~57, Apr. 2025.

\bibitem{CapodieciCBP18}
N.~Capodieci, R.~Cavicchioli, M.~Bertogna, and A.~Paramakuru, ``Deadline-based scheduling for {GPU} with preemption support,'' in {\em 2018 {IEEE} Real-Time Systems Symposium, {RTSS} 2018, Nashville, TN, USA, December 11-14, 2018}, pp.~119--130, {IEEE} Computer Society, 2018.

\bibitem{OlmedoCMMB20}
I.~S. Olmedo, N.~Capodieci, J.~L. Martinez, A.~Marongiu, and M.~Bertogna, ``Dissecting the {CUDA} scheduling hierarchy: a performance and predictability perspective,'' in {\em {IEEE} Real-Time and Embedded Technology and Applications Symposium, {RTAS} 2020, Sydney, Australia, April 21-24, 2020}, pp.~213--225, {IEEE}, 2020.

\bibitem{YaoZSLLSA18}
S.~Yao, Y.~Zhao, H.~Shao, S.~Liu, D.~Liu, L.~Su, and T.~F. Abdelzaher, ``Fastdeepiot: Towards understanding and optimizing neural network execution time on mobile and embedded devices,'' in {\em Proceedings of the 16th {ACM} Conference on Embedded Networked Sensor Systems, SenSys 2018, Shenzhen, China, November 4-7, 2018} (G.~S. Ramachandran and B.~Krishnamachari, eds.), pp.~278--291, {ACM}, 2018.

\bibitem{ZhuZL18}
X.~Zhu, W.~Zhou, and H.~Li, ``Adaptive layerwise quantization for deep neural network compression,'' in {\em 2018 {IEEE} International Conference on Multimedia and Expo, {ICME} 2018, San Diego, CA, USA, July 23-27, 2018}, pp.~1--6, {IEEE} Computer Society, 2018.

\bibitem{Yao0LWLSA20}
S.~Yao, J.~Li, D.~Liu, T.~Wang, S.~Liu, H.~Shao, and T.~F. Abdelzaher, ``Deep compressive offloading: speeding up neural network inference by trading edge computation for network latency,'' in {\em SenSys '20: The 18th {ACM} Conference on Embedded Networked Sensor Systems, Virtual Event, Japan, November 16-19, 2020} (J.~Nakazawa and P.~Huang, eds.), pp.~476--488, {ACM}, 2020.

\bibitem{ZhangWXD24}
B.~Zhang, T.~Wang, S.~Xu, and D.~S. Doermann, {\em Neural Networks with Model Compression}.
\newblock Springer, 2024.

\bibitem{LiCLZTL25}
S.~Li, J.~Chen, S.~Liu, C.~Zhu, G.~Tian, and Y.~Liu, ``{MCMC:} multi-constrained model compression via one-stage envelope reinforcement learning,'' {\em {IEEE} Trans. Neural Networks Learn. Syst.}, vol.~36, no.~2, pp.~3410--3422, 2025.

\bibitem{DaiFMH25}
W.~Dai, J.~Fan, Y.~Miao, and K.~Hwang, ``Deep learning model compression with rank reduction in tensor decomposition,'' {\em {IEEE} Trans. Neural Networks Learn. Syst.}, vol.~36, no.~1, pp.~1315--1328, 2025.

\bibitem{DBLP:conf/rtss/Bateni018}
S.~Bateni and C.~Liu, ``Apnet: Approximation-aware real-time neural network,'' in {\em 2018 {IEEE} Real-Time Systems Symposium, {RTSS} 2018, Nashville, TN, USA, December 11-14, 2018}, pp.~67--79, {IEEE} Computer Society, 2018.

\bibitem{DBLP:conf/rtas/LeeN20}
S.~Lee and S.~Nirjon, ``Subflow: {A} dynamic induced-subgraph strategy toward real-time {DNN} inference and training,'' in {\em {IEEE} Real-Time and Embedded Technology and Applications Symposium, {RTAS} 2020, Sydney, Australia, April 21-24, 2020}, pp.~15--29, {IEEE}, 2020.

\bibitem{3724126}
M.~F. Babar and M.~Hasan, ``Optimizing confidential deep learning for real-time systems,'' {\em ACM Trans. Cyber-Phys. Syst.}, Mar. 2025.

\bibitem{YaoHZSLLW0A20}
S.~Yao, Y.~Hao, Y.~Zhao, H.~Shao, D.~Liu, S.~Liu, T.~Wang, J.~Li, and T.~F. Abdelzaher, ``Scheduling real-time deep learning services as imprecise computations,'' in {\em 26th {IEEE} International Conference on Embedded and Real-Time Computing Systems and Applications, {RTCSA} 2020, Gangnueng, Korea (South), August 19-21, 2020}, pp.~1--10, {IEEE}, 2020.

\bibitem{XuLFL0C24}
Y.~Xu, Z.~Liu, X.~Fu, S.~Liu, F.~Wu, and G.~Chen, ``{FLEX:} adaptive task batch scheduling with elastic fusion in multi-modal multi-view machine perception,'' in {\em {IEEE} Real-Time Systems Symposium, {RTSS} 2024, York, UK, December 10-13, 2024}, pp.~294--307, {IEEE}, 2024.

\bibitem{SoyyigitYY24}
A.~Soyyigit, S.~Yao, and H.~Yun, ``{VALO:} {A} versatile anytime framework for lidar-based object detection deep neural networks,'' {\em {IEEE} Trans. Comput. Aided Des. Integr. Circuits Syst.}, vol.~43, no.~11, pp.~4045--4056, 2024.

\bibitem{LiuYFTYBYSA2022}
S.~Liu, S.~Yao, X.~Fu, H.~Shao, R.~Tabish, S.~Yu, A.~Bansal, H.~Yun, L.~Sha, and T.~F. Abdelzaher, ``Real-time task scheduling for machine perception in intelligent cyber-physical systems,'' {\em {IEEE} Trans. Computers}, vol.~71, no.~8, pp.~1770--1783, 2022.

\bibitem{HuLAWD21}
Y.~Hu, S.~Liu, T.~F. Abdelzaher, M.~B. Wigness, and P.~David, ``On exploring image resizing for optimizing criticality-based machine perception,'' in {\em 27th {IEEE} International Conference on Embedded and Real-Time Computing Systems and Applications, {RTCSA} 2021, Houston, TX, USA, August 18-20, 2021}, pp.~169--178, {IEEE}, 2021.

\bibitem{LiuYFTYBYSA2020}
S.~Liu, S.~Yao, X.~Fu, R.~Tabish, S.~Yu, A.~Bansal, H.~Yun, L.~Sha, and T.~F. Abdelzaher, ``On removing algorithmic priority inversion from mission-critical machine inference pipelines,'' in {\em 41st {IEEE} Real-Time Systems Symposium, {RTSS} 2020, Houston, TX, USA, December 1-4, 2020}, pp.~319--332, {IEEE}, 2020.

\bibitem{YaoZZSA17}
S.~Yao, Y.~Zhao, A.~Zhang, L.~Su, and T.~F. Abdelzaher, ``Deepiot: Compressing deep neural network structures for sensing systems with a compressor-critic framework,'' in {\em Proceedings of the 15th {ACM} Conference on Embedded Network Sensor Systems, SenSys 2017, Delft, Netherlands, November 06-08, 2017} (M.~R. Eskicioglu, ed.), pp.~4:1--4:14, {ACM}, 2017.

\bibitem{teerapittayanon2016branchynet}
S.~Teerapittayanon, B.~McDanel, and H.~T. Kung, ``Branchynet: Fast inference via early exiting from deep neural networks,'' in {\em 23rd International Conference on Pattern Recognition (ICPR)}, pp.~2464--2469, 2016.

\bibitem{Ila_Gokarn_2024_1}
I.~Gokarn, H.~Sabbella, Y.~Hu, T.~F. Abdelzaher, and A.~Misra, ``Demonstrating canvas-based processing of multiple camera streams at the edge,'' in {\em 16th International Conference on COMmunication Systems {\&} NETworkS, {COMSNETS} 2024, Bengaluru, India, January 3-7, 2024}, pp.~297--299, {IEEE}, 2024.

\bibitem{Ila_Gokarn_2024_2}
I.~Gokarn, ``Criticality aware canvas-based visual perception at the edge,'' in {\em Proceedings of the 22nd Annual International Conference on Mobile Systems, Applications and Services, {MOBISYS} 2024, Minato-ku, Tokyo, Japan, June 3-7, 2024} (T.~Okoshi, J.~Ko, and R.~LiKamWa, eds.), pp.~751--753, {ACM}, 2024.

\end{thebibliography}

\end{document}